\documentclass{article}
\usepackage{amssymb}
\usepackage{graphicx}
\usepackage{booktabs}  
\usepackage{multirow}  
\usepackage{subcaption}
\usepackage{todonotes}
\usepackage{wrapfig}
\usepackage{tcolorbox}
\usepackage{subcaption}
\usepackage{xspace}

\usepackage[preprint]{corl_2025} 

\usepackage{calc}
\usepackage{makecell}
\usepackage{array}

\newcolumntype{L}[1]{>{\raggedright\arraybackslash}m{#1}}

\newcommand{\addAugFig}[2]{\includegraphics[width=\dimexpr#1\linewidth\relax]{assets/#2}}

\newcommand{\ourmethod}{\mbox{\textsc{DreamGen}}\xspace}
\newcommand{\ourbench}{\texttt{{\mbox{\textsc{DreamGen Bench}}}}\xspace}

\title{\ourmethod: Unlocking Generalization in \\Robot Learning through Video World Models}

\usepackage{authblk}  
\usepackage{hyperref}
\usepackage{footmisc}

\newcommand{\symbolfootnotetext}[2]{%
  \begingroup%
  \renewcommand{\thefootnote}{#1}%
  \let\oldfootnoterule\footnoterule%
  \renewcommand{\footnoterule}{\oldfootnoterule}%
  \footnotetext{#2}%
  \endgroup%
}

\author{
  \textbf{Joel Jang}\textsuperscript{1,2,*}
  \textbf{Seonghyeon Ye}\textsuperscript{1,3,*}
  \textbf{Zongyu Lin}\textsuperscript{1,4,*}
  \textbf{Jiannan Xiang}\textsuperscript{1,5,*}
  \vspace{-2ex} \\
  \textbf{Johan Bjorck}\textsuperscript{1} 
  \textbf{Yu Fang}\textsuperscript{1}
  \textbf{Fengyuan Hu}\textsuperscript{1}
  \textbf{Spencer Huang}\textsuperscript{1}
  \textbf{Kaushil Kundalia}\textsuperscript{1}
  \textbf{Lin Yen-Chen}\textsuperscript{1}
  \textbf{Loic Magne}\textsuperscript{1}
  \vspace{0.25ex} \\
  \textbf{Ajay Mandlekar}\textsuperscript{1}
  \textbf{Avnish Narayan}\textsuperscript{1}
  \textbf{You Liang Tan}\textsuperscript{1} 
  \textbf{Guanzhi Wang}\textsuperscript{1,6} 
  \textbf{Jing Wang}\textsuperscript{1,7}
  \textbf{Qi Wang}\textsuperscript{1}
  \textbf{Yinzhen Xu}\textsuperscript{1}
  \vspace{0.25ex} \\
  \textbf{Xiaohui Zeng}\textsuperscript{1}
  \textbf{Kaiyuan Zheng}\textsuperscript{2} 
  \textbf{Ruijie Zheng}\textsuperscript{1,8}
  \vspace{0.5ex} \\
  \textbf{Ming-Yu Liu}\textsuperscript{1}
  \textbf{Luke Zettlemoyer}\textsuperscript{2}
  \textbf{Dieter Fox}\textsuperscript{1,2}
  \textbf{Jan Kautz}\textsuperscript{1}
  \textbf{Scott Reed}\textsuperscript{1,$\dagger$}
  \textbf{Yuke Zhu}\textsuperscript{1,9,$\dagger$}
  \textbf{Linxi Fan}\textsuperscript{1,$\dagger$}
  \vspace{1ex} \\ 
  {\textsuperscript{1}NVIDIA}\quad{\textsuperscript{2}University of Washington}\quad{\textsuperscript{3}KAIST}\quad{\textsuperscript{4}UCLA}\quad{\textsuperscript{5}UCSD}\\
  {\textsuperscript{6}CalTech}\quad{\textsuperscript{7}NTU}\quad{\textsuperscript{8}University of Maryland}\quad{\textsuperscript{9}UT Austin}
  \vspace{1ex} \\ 
  {\small \url{https://research.nvidia.com/labs/gear/dreamgen}}
}

\begin{document}

\maketitle

\renewcommand{\thefootnote}{\fnsymbol{footnote}}

\setcounter{footnote}{0}

\symbolfootnotetext{*}{Equal contribution.}
\symbolfootnotetext{$\dagger$}{Equal advising.}

\renewcommand{\thefootnote}{\arabic{footnote}}
\setcounter{footnote}{0}
\vspace{-6ex}
\begin{figure}[h!]
\centering
\includegraphics[width=0.9\textwidth]{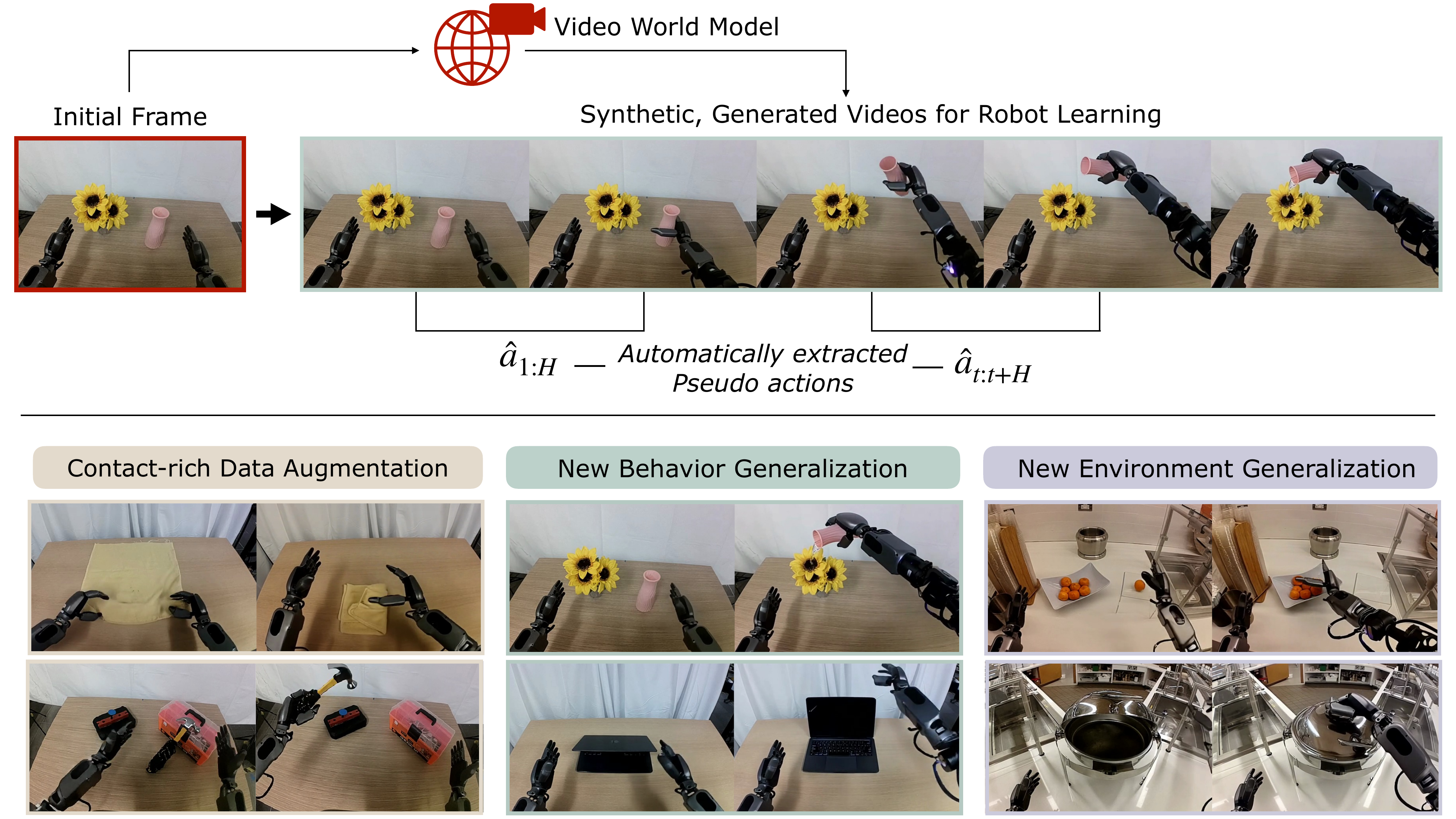}
\caption{\textbf{Generalization through \ourmethod}. We enable 2D visuomotor robot policies to generalize to \textbf{new environments} with \textbf{new behaviors}, while \textit{only} collecting teleoperation data for a \textbf{single} behavior type (pick\&place) in a \textbf{single} environment by utilizing video world models as synthetic data generators.}
\label{fig:main_overview}
\end{figure}



\begin{abstract}
    We introduce \ourmethod, a simple yet highly effective 4-stage pipeline for training robot policies that generalize across behaviors and environments through \textit{neural trajectories}—synthetic robot data generated from video world models. \ourmethod leverages state-of-the-art image-to-video generative models, adapting them to the target robot embodiment to produce photorealistic synthetic videos of familiar or novel tasks in diverse environments. Since these models generate only videos, we recover pseudo-action sequences using either a latent action model or an inverse-dynamics model (IDM). Despite its simplicity, \ourmethod unlocks strong behavior and environment generalization: a humanoid robot can perform 22 new behaviors in both seen and unseen environments, while requiring teleoperation data from only a single pick-and-place task in one environment. To evaluate the pipeline systematically, we introduce \ourbench, a video generation benchmark that shows a strong correlation between benchmark performance and downstream policy success. Our work establishes a promising new axis for scaling robot learning well beyond manual data collection.
\end{abstract}



\section{Introduction}

Robot foundation models trained on large-scale human teleoperation data have shown strong potential for general-purpose robotic systems to perform dexterous real-world tasks~\citep{rt22023arxiv, black2410pi0, team2025gemini, bu2025agibot, bjorck2025gr00t, intelligence2025pi}. However, this paradigm relies heavily on collecting teleoperation data manually for every new task and environment, which remains costly and labor-intensive. Synthetic data generation in simulation offers an appealing alternative, but it often requires significant manual engineering and suffers from sim2real gap when deploying visuomotor policies on physical robots. To address these challenges, we propose \ourmethod, a new synthetic data pipeline that leverages video world models to create realistic training data at scale with minimal manual labor or engineering.
\begin{figure}[t!]
\centering
\includegraphics[width=0.99\textwidth]{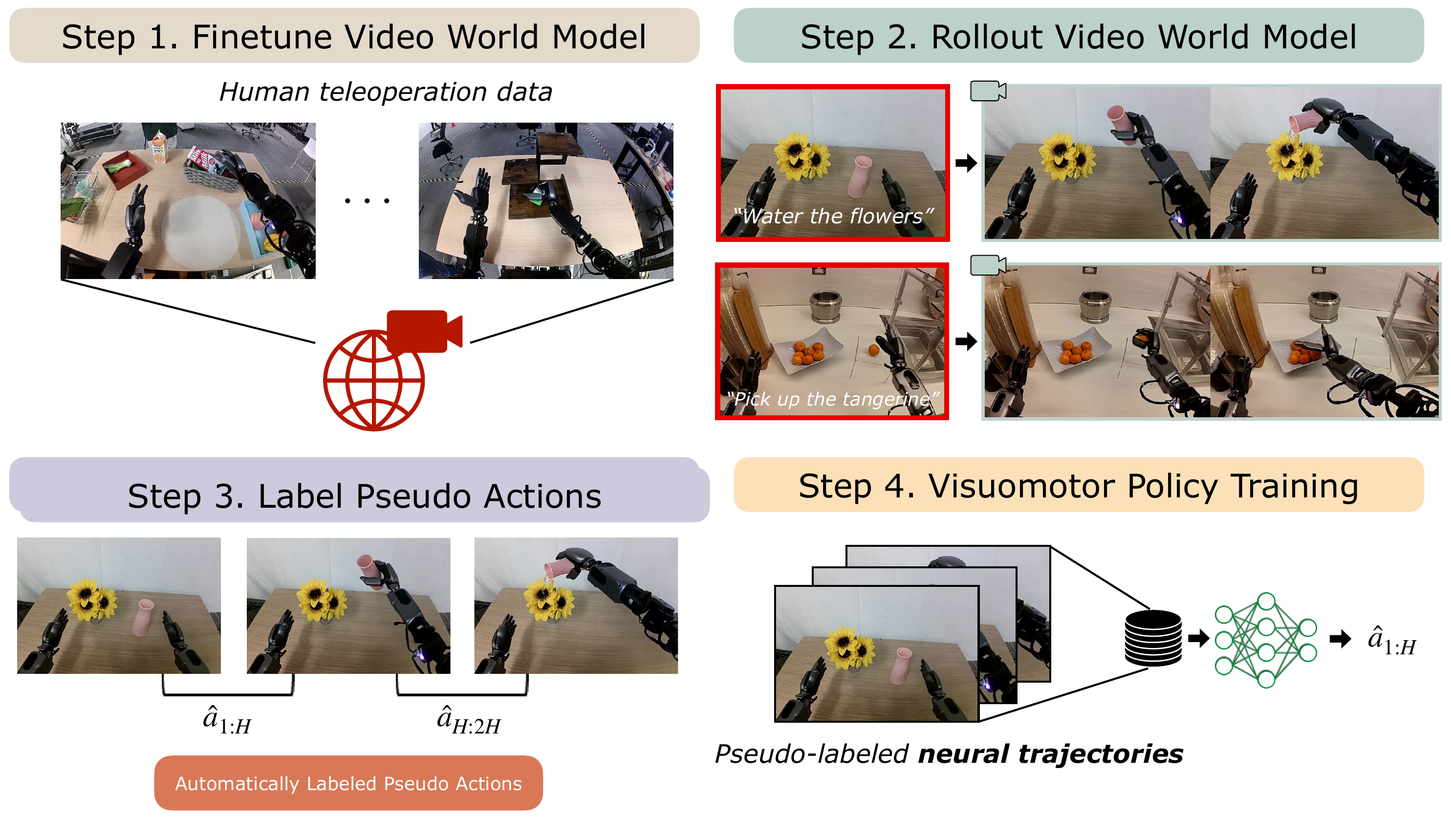}

\caption{\textbf{\ourmethod Overview.} We begin by fine-tuning a video world model on teleoperated robot trajectories. Given an initial frame and a language instruction, the model generates video rollouts depicting the intended behavior. As these videos lack action annotations, we infer pseudo-actions using either a latent action model or an inverse dynamics model, forming what we call \textit{neural trajectories}. Finally, we train visuomotor robot policies on these neural trajectories.
\label{fig:groot_stages}}
\end{figure}


\ourmethod follows a simple 4-step recipe (Figure~\ref{fig:groot_stages}) for applying state-of-the-art video generative models~\citep{agarwal2025cosmos, yang2024cogvideox, wang2025wan, kong2024hunyuanvideo, lin2024stiv, xiang2024pandora}, also known as \textit{video world models}, to generate synthetic training data. This pipeline is designed to be general-purpose across different robots, environments, and tasks. (1) We fine-tune video world models on a target robot to capture the dynamics and kinematics of the specific embodiment; (2) we prompt the model with pairs of initial frames and language instructions to generate large volumes of robot videos, capturing both familiar behaviors from fine-tuning and novel ones in unseen settings; (3) we then extract pseudo-actions using either a latent action model~\citep{ye2025latent} or an inverse dynamics model (IDM)\citep{baker2022video}; (4) finally, we use the resulting video-action sequence pairs, dubbed \textit{neural trajectories}, for training downstream visuomotor policies. While prior work has focused on using video world models as real-time planners~\citep{du2023learning, zhou2024robodreamer, ko2024learning, yang2024learning, du2024video}, \ourmethod instead treats them as \textit{synthetic data generators}, unlocking their strong priors for physical reasoning, naturalistic motion, and language grounding.


First, we investigate \ourmethod for generating additional training data for tasks where teleoperation data is already available, both in simulation and the real world. In simulation, we apply \ourmethod to the RoboCasa benchmark~\citep{robocasa2024}, scaling synthetic data up to 333$\times$ relative to the original human demonstrations. This yields log-linear improvements in policy performance as the number of neural trajectories increases (Figure~\ref{fig:24P_sim_mg}). In the real world, we validate our approach on 9 diverse tasks on Fourier GR1, Franka Emika, and SO-100 robots, demonstrating the flexibility of our pipeline across embodiments and challenging dexterous tasks that are difficult to simulate, such as folding towels, wiping liquids, using hammers, and scooping M\&Ms. \ourmethod show consistent gains on success rate across all robots: from 37\% to 46.4\% on average of 4 GR1 humanoid tasks, 23\% to 37\% on average of 3 Franka tasks, and from 21\% to 45.5\% on average of 2 SO-100 tasks, all using just 10 to 13 real-world trajectories per task.


Next, we highlight two key generalization capabilities unlocked by \ourmethod: \textbf{behavior generalization} and \textbf{environment generalization}. For behavior generalization, we enable the GR1 humanoid to perform 22 novel behaviors, such as pouring, opening/closing articulated objects, and manipulating a variety of tools. Note that the original teleoperation dataset only includes pick-and-place and no other verbs. For environment generalization, we prompt video world models (fine-tuned on just a single environment) with initial frames from 10 new environments. This allows us to train visuomotor policies that generalize to novel behaviors and settings using only teleoperation data from a single task in a single environment. These represent true zero-to-one improvements – GR00T N1 trained on pick-and-place alone achieves 0\% success rates on most novel behavior and environment experiments, while \ourmethod enables 43.2\% success rates on new behaviors in seen environments and 28.5\% in completely unseen environments. These empirical results point towards a new paradigm for scalable robot learning without extensive manual demonstrations.


Lastly, we introduce \ourbench (Section~\ref{sec:wm_benchmark}), a new video generation benchmark designed to evaluate how well different video world models adapt to novel robot embodiments. We assess whether 8 models, 4 zero-shot and 4 fine-tuned, can generate robot videos that involve manipulating unseen objects, performing unseen behaviors, and operating in unseen environments, all while abiding by the laws of physics. Empirically, we find that models with higher scores also yield stronger downstream robot policy performance. \ourbench provides a diagnostic and low-cost way to connect video world models to robotics, without requiring a physical robot in the loop. We hope this offers an accessible pathway for video model researchers to contribute to robot learning.

\vspace{-1.5mm}
\section{\ourmethod}
\label{sec:method}
\vspace{-2.5mm}

In the next subsections, we describe in detail the 4 different steps (shown in Figure \ref{fig:groot_stages}) of \ourmethod, creating and utilizing neural trajectories to train visuomotor robot policies.

\subsection{Video World Model Fine-tuning}
In the initial phase, we fine-tune video world models on human-teleoperated robot trajectories. This adaptation enables the model to learn the robot's physical constraints and movement capabilities. To mitigate forgetting prior internet video knowledge, we use Low-Rank Adaptation (LoRA)~\citep{hu2022lora} by default for the different video world model fine-tuning we conduct.
When fine-tuning these models, we look at two metrics, \textit{instruction following} and \textit{physics following}, to determine whether the video world model has been optimally adapted to the target robot domain (details provided in Section \ref{sec:wm_benchmark}). For the majority of our downstream robot experiments, we utilize WAN2.1~\citep{wang2025wan} as our base video world model. In cases where there are multiple viewpoints in the training dataset (RoboCasa~\citep{robocasa2024} and DROID~\citep{khazatsky2024droid}), we concatenate the viewpoints into a 2$\times$2 grid (with one grid with black pixels) and fine-tune the video world models.\footnote{Examples are shown in Appendix \ref{appen:multiview_examples}.} We also observe that the optimal amount of fine-tuning required for each video world model and fine-tuning data pair differs.\footnote{We provide the hyperparameters (learning rate, number of epochs, etc.) used for all of the experimental setups in Appendix \ref{appen:wm_hyperparameter}.}

\subsection{Video World Model Rollout}
After fine-tuning the video world models on the target robot embodiment, we generate synthetic robot videos using various initial frames and language instructions. 
For simulation experiments, we collect new initial frames from the simulator, randomizing the locations of the target objects or environments for each task. For real-world experiments, we manually take new initial frames while randomizing the location of the target object. For environment generalization experiments, we also take initial frames of new environments, while we restrict ourselves to training the video world model collected from a single environment (pictures shown in Appendix \ref{appen:env}). Lastly, we manually come up with novel behavior prompts for the behavior generalization experiments, and also include all of the candidates in our video benchmark 
in Section \ref{sec:wm_benchmark}.\footnote{Even though collecting new initial frames requires some manual work, it significantly alleviates the need for collecting new teleoperation data. Furthermore, we hope to utilize image-to-image diffusion techniques to alleviate this burden, where we can start off with a single initial frame, and randomize new initial frames by impainting the object locations, type of objects, as well as the environment for future work.}

\subsection{Pseudo Action Labeling}
\label{subsec:pseudo_actions}
\begin{figure}[t!]
    \centering
    \begin{subfigure}[b]{0.46\textwidth}
        \centering
        \includegraphics[width=\textwidth]{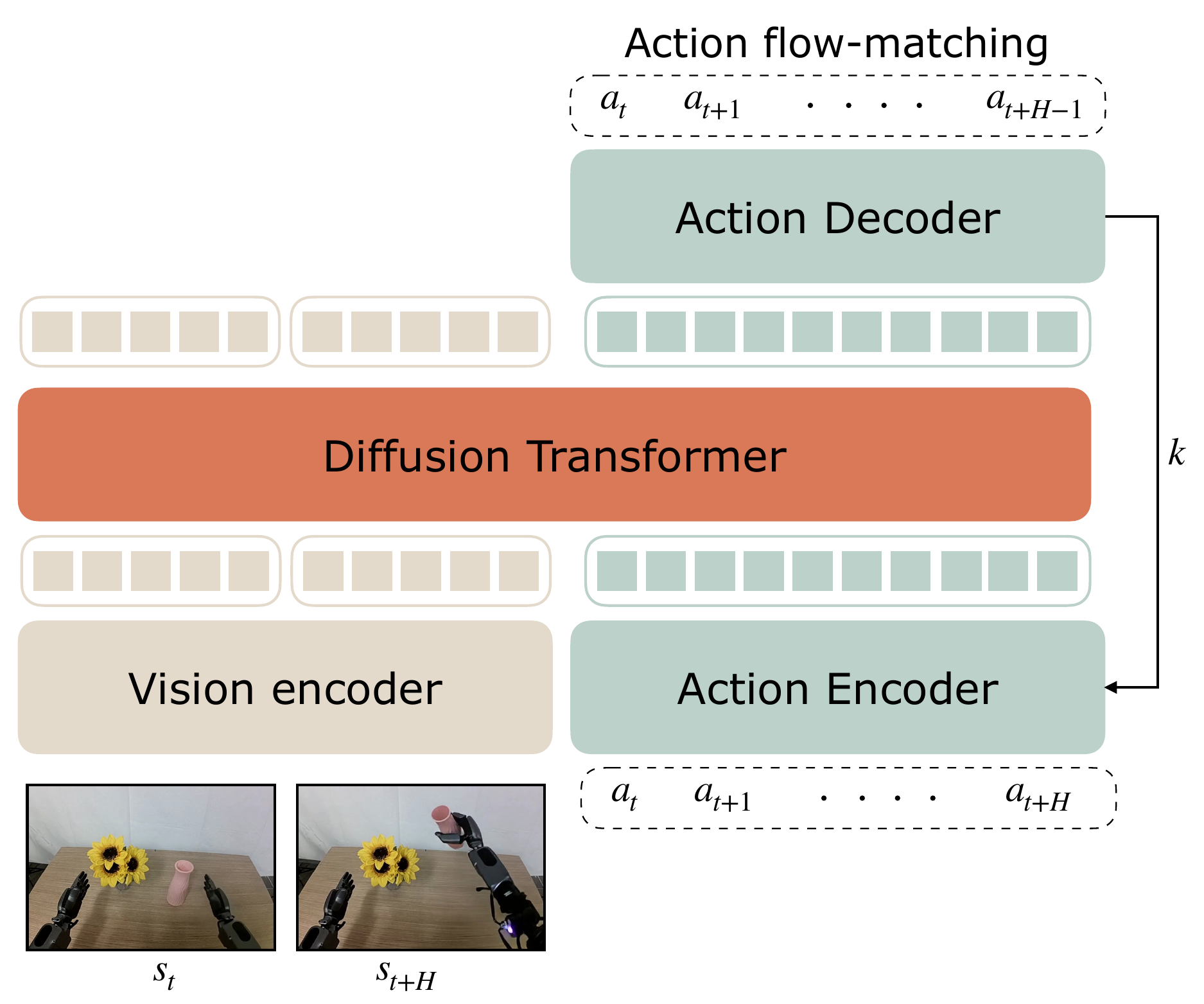}
        \caption{Inverse Dynamics Model (IDM)~\citep{baker2022videopretrainingvptlearning}}
        \label{fig:subfig1}
    \end{subfigure}
    \hspace{9mm}
    \begin{subfigure}[b]{0.23\textwidth}
        \centering
        \includegraphics[width=\textwidth]{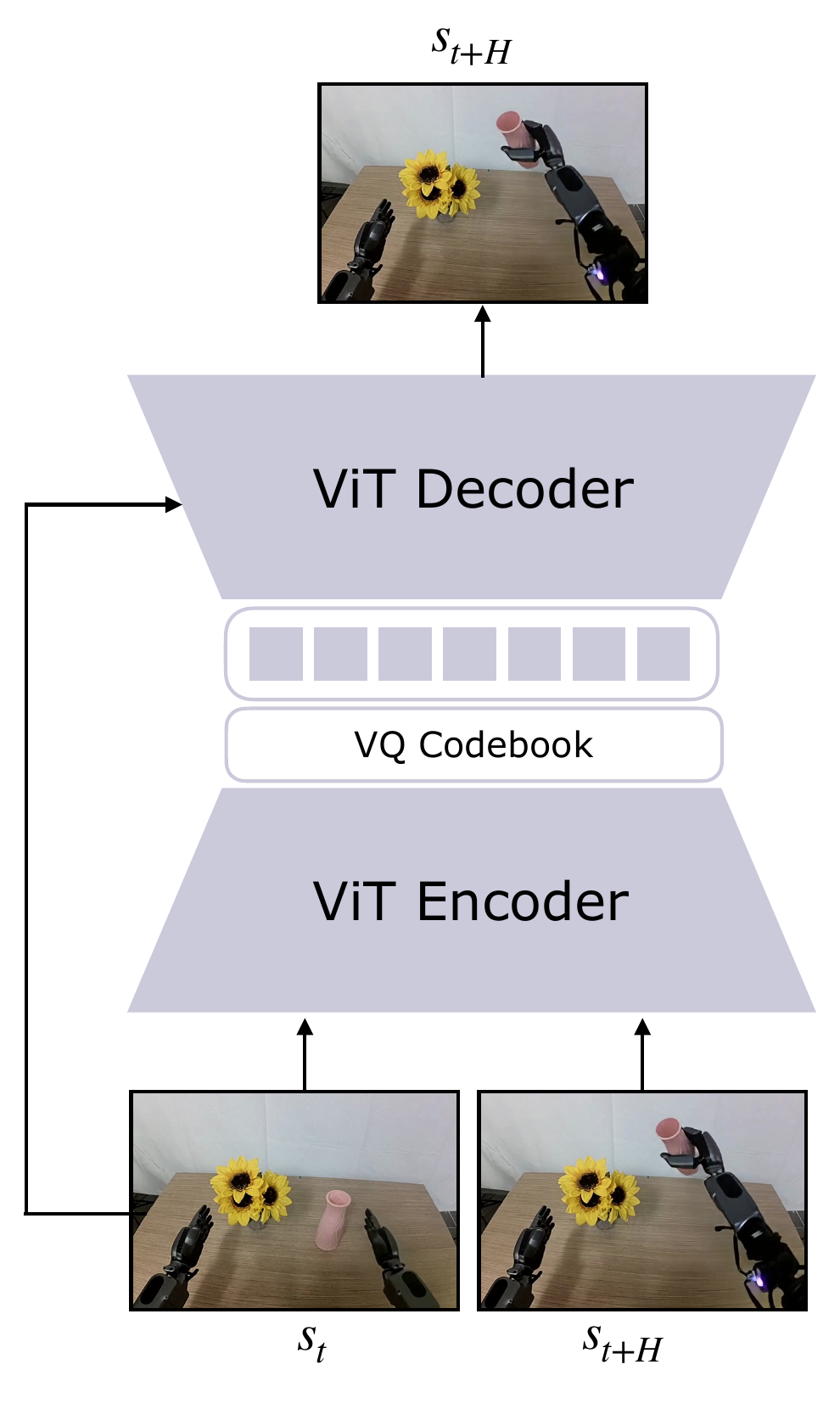}
        \caption{LAPA~\citep{ye2025latent}}
        \label{fig:subfig2}
    \end{subfigure}
    \caption{\textbf{Extracting Pseudo Actions.} (a) shows the architecture of our IDM model and (b) shows the architecture of our latent action model.}
    \label{fig:pseudo_actions}
\end{figure}

Figure \ref{fig:pseudo_actions} shows the (a) architecture we use to train the IDM model and the (b) architecture that we use to train the latent action model (LAPA), both used to extract pseudo action labels for the generated videos. 

\paragraph{IDM Actions.} For the inverse dynamics model (IDM) architecture, we use diffusion transformers with SigLIP-2 vision encoder and train with a flow matching objective. IDM is conditioned on two image frames and is trained to predict action chunks between the image frames (Figure \ref{fig:pseudo_actions}). We do not explicitly use any language or proprioception as input, since we want the IDM model to only capture the dynamics of the robot. For the IDM training data, we use the same dataset used to train the video world models for each setup, unless explicitly stated otherwise. After training, we employ a sliding window approach for pseudo-labeling: the IDM predicts $H$ actions, $\hat{a}_{t}$ to $\hat{a}_{t+H}$. Next, it slides one window and predicts another $H$ actions, $\hat{a}_{t+1}$ to $\hat{a}_{t+1+H}$, and so forth. More details are provided in Appendix \ref{appen:pseudo_actions}.

\paragraph{Latent Actions.} For latent actions, we use the LAPA latent action model~\citep{ye2025latent}, which has a transformer encoder-decoder architecture and is trained on diverse robot and human videos. The latent action model is trained with a VQ-VAE objective so that the latent actions can capture the visual delta information between two frames in a video. To obtain the latent actions from the generated videos, we condition the latent action model on the current frame and the future frame (1 second ahead) of the trajectory. We use the pre-quantized continuous embedding as the latent action following GR00T N1~\citep{bjorck2025gr00t}. The exact training data mixture used to train the latent action model is provided in Table \ref{tab:dataset_stats}. One benefit of latent actions is that it does not require actually having ground-truth actions for the target robot embodiment when training latent action models.

\subsection{Policy Training on Neural Trajectories}
Lastly, we train visuomotor robot policies on neural trajectories generated by \ourmethod by conditioning on language instruction and image observations. We condition state information with zero values, since neural trajectories do not contain state information.\footnote{From preliminary experiments, we observed that having zero state does not harm the performance. We leave training the IDM to predict state information for future work.} More specifically, given $o_{t}$, the image observation, and $i_{t}$, the task instruction, we train the policies to generate $\hat{a}_{t:t+H}$, which can be either latent actions or IDM-labeled actions from the previous subsection. Since neural trajectories are independent of the underlying robot policy architecture, we showcase the effectiveness of \ourmethod for generating synthetic training data for 3 different visuomotor policy models, Diffusion Policy~\citep{chi2023diffusion}, $\pi_0$~\citep{black2410pi0}, and GR00T N1~\citep{bjorck2025gr00t}. 

We propose two scenarios of training with neural trajectories: co-training with real-world trajectories, and solely training on the neural trajectories labeled with IDM actions. When we co-train neural trajectories with real trajectories, we co-train with a sampling ratio of 1:1. For GR00T N1, we treat the two types of trajectories as separate embodiments by using separate action encoder and decoder. For behavior and environment generalization experiments, we only use neural trajectories for policy training. 

\section{Experiments}

In this section, we demonstrate three key applications of \ourmethod: (1) Augmenting training data for existing tasks, (2) Enabling generalization to novel behaviors, and (3) Enabling generalization to novel environments. 

\subsection{Training Data Augmentation}
\label{section:training_data}
\begin{figure}[t!]
    \centering
    \includegraphics[width=0.95\textwidth]{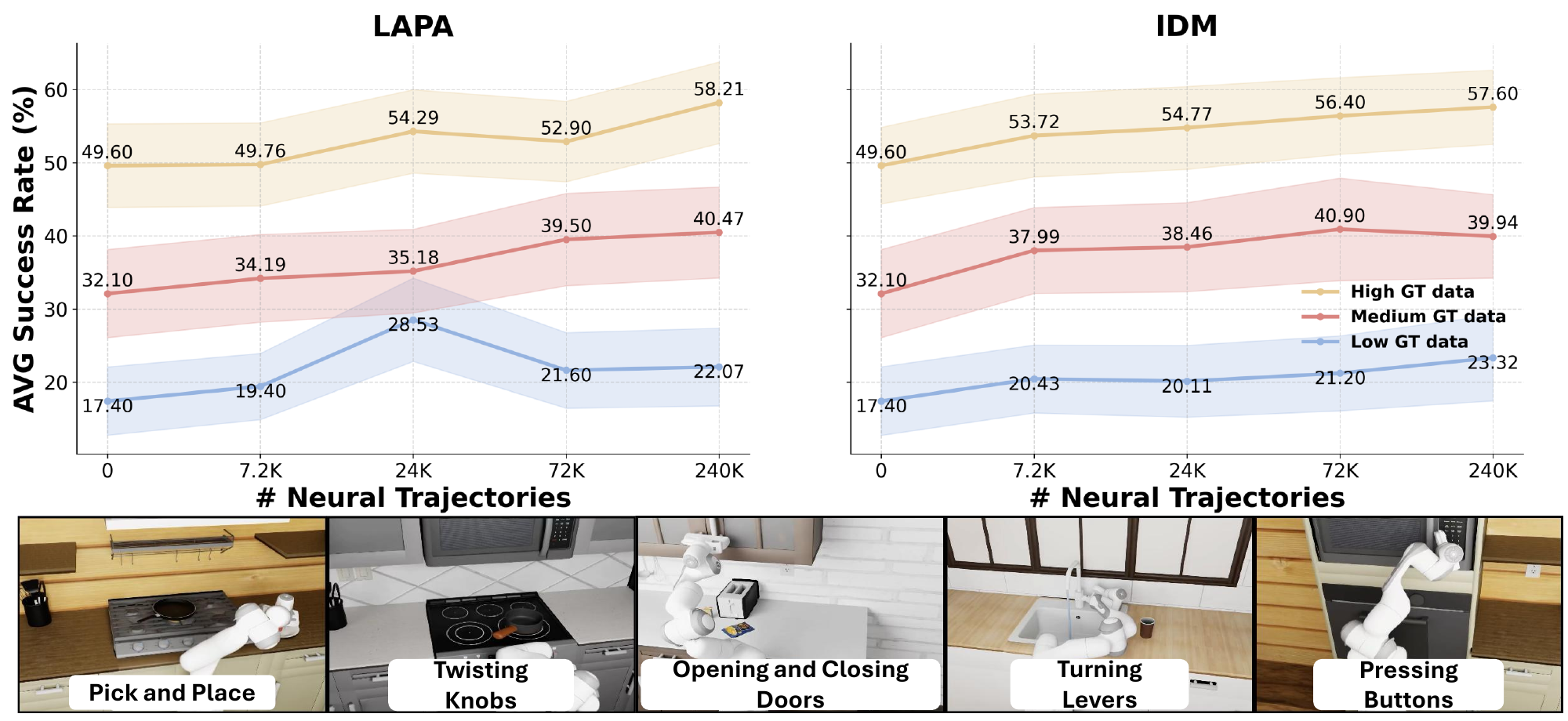}
    \caption{\textbf{Scaling \# of Neural Trajectories in RoboCasa}. We vary the sizes of neural trajectories (x-axis) and ground-truth trajectories (low, mid, high) and report results with both latent and IDM actions as pseudo action labels. We report the average success rate (\%) across 24 tasks. The results at $x=0$ correspond to the baseline \textbf{only} trained on ground-truth videos.}
    \label{fig:24P_sim_mg}
\end{figure}

\begin{figure}[ht!]
    \centering
    \includegraphics[width=\columnwidth]{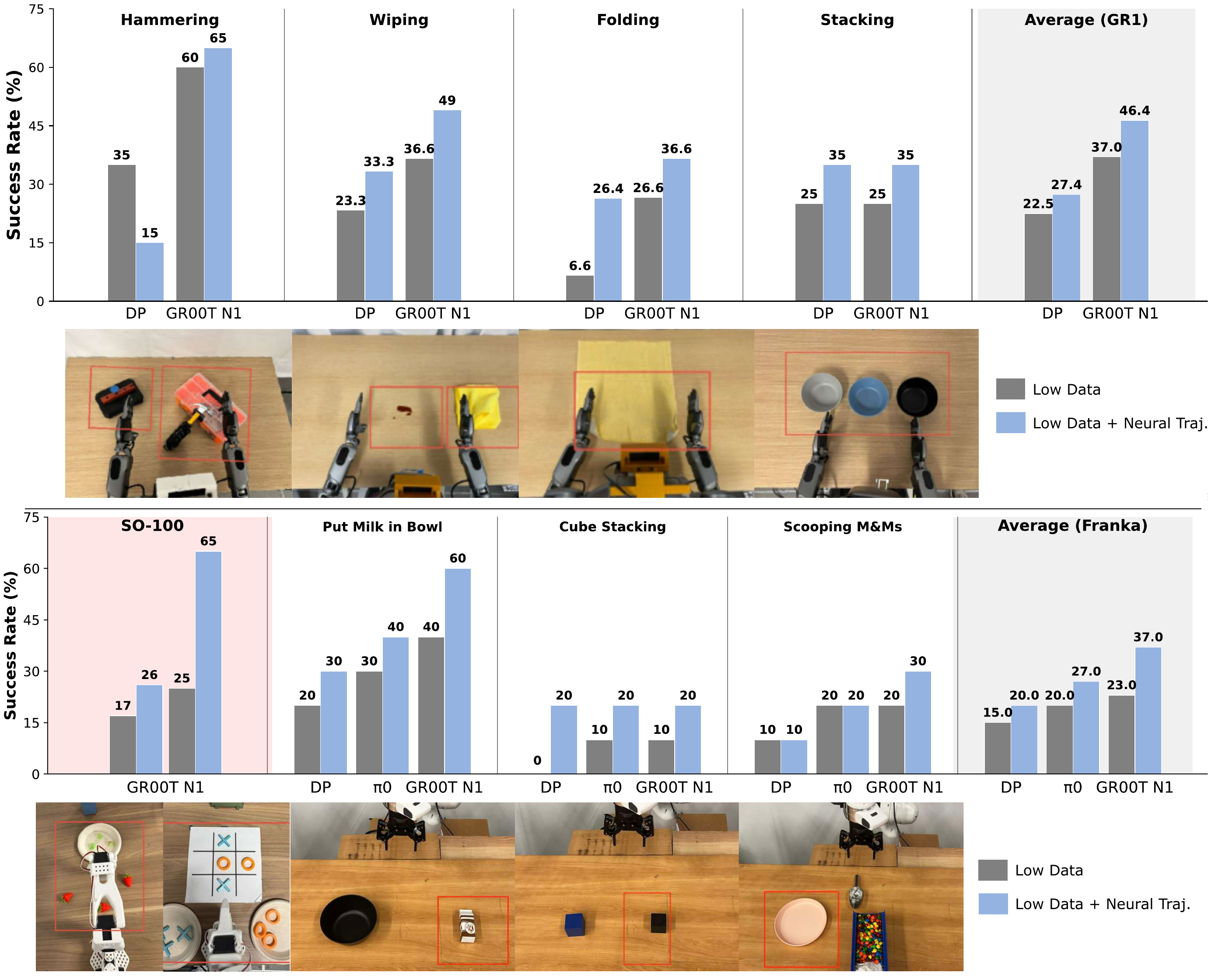}
    \caption{\textbf{Real-world Robot Evaluation Results}. The red rectangular box shows the range of object randomization during training and evaluation. \textit{Low Data} denotes training 10\% of available training data (only 10 trajectories per task except for GR1-folding, where we used 25 trajectories), and \textit{Low Data + Neural Traj.} denotes co-training with neural trajectories.}
    \label{fig:real_robot_mg}
    
\end{figure}

For simulation experiments, we evaluate our pipeline on the RoboCasa benchmark~\citep{robocasa2024}, using the same training and evaluation protocol as outlined in the original work. For real-world experiments, we evaluate on 9 real-world tasks across three embodiments: the GR1 humanoid robot, the Franka arm robot, and the low-cost SO-100 robot arm.

\paragraph{Simulation experiments} 
Figure \ref{fig:24P_sim_mg} shows the downstream robot policy results as we scale the total number of neural trajectories in three different scenarios of ground-truth data: low data (720), mid-data (2.4k), and high-data (7.2k) on RoboCasa. Each scenario determines how \textit{strong} our IDM model can become, since the more ground-truth data we have about a given robot, the more useful dynamics the model can learn. In this particular setup, we train our video world model on 1,200 original human demonstrations, whereas IDM and policy training are conducted in different data scenarios from the benchmark.\footnote{RoboCasa Benchmark consists of three different viewpoints for visuomotor policy training: left, right, and wrist. We utilize GR00T N1~\citep{bjorck2025gr00t} as the base robot policy for this experiment.}

First, we observe that co-training with neural trajectories yields a performance boost for both IDM and LAPA actions across all data regime scenarios. Since both approaches have similar effects, we use IDM as the default for the rest of the experiments, as IDM actions enable \textit{solely} training on neural trajectories and evaluating the policy performance, and in all of our experimental set-up, we do have access to teleoperation data to train strong enough IDMs for each robot embodiment.~\footnote{Enabling zero-shot generalization to novel behaviors and novel environments with robot embodiments with \textit{zero} ground-truth data still remains an open research question.} Second, we observe that there is a consistent log-linear slope between the total number of neural trajectories and the downstream robot policy performance. This hints towards a potential for a new paradigm in robot learning, as synthetic data generation through neural trajectories is significantly more scalable compared to the traditional method of manual teleoperation for imitation learning. Lastly, we show that \textit{solely} training on neural trajectories with IDM actions enables us to reach a non-trivial performance (20.6\% average success rate across 24 tasks), further highlighting the quality of neural trajectories (a detailed breakdown of results is provided in Appendix \ref{appen:robocasa_detail_result}).

\paragraph{Real-world Experiments} 
For real-world experiments, we collect 100 trajectories per task for the four GR1 and three Franka tasks. For the two SO-100 tasks, we collect 40 and 50 trajectories for the strawberry pick-and-place and tic-tac-toe tasks, respectively. Details of the data collection and evaluation criteria for each of the 9 tasks are provided in the Appendix \ref{appendix:robot-evals}, and details of the video world model training procedure for each task are provided in Appendix \ref{appen:video_world_model}. As default, we use only 10\% of the collected trajectories for our main experiment to test data efficiency for GR1 and Franka tasks (only 10 real-world trajectories per task) and 25\% of the collected trajectories for SO-100 tasks (10 and 13 trajectories per task).\footnote{We also provide the evaluation results of models trained on ``High Data'' (100\% of training data) in Appendix \ref{appen:full_realworld_results}.} We generate 300 neural trajectories for each GR1 task, 100 neural trajectories for each Franka task, and 40 and 50 neural trajectories for the two SO-100 tasks, respectively, to co-train with real-world trajectories with a 1:1 sampling ratio.

As shown in Figure \ref{fig:real_robot_mg}, neural trajectories consistently improve performance for different visuomotor policies (Diffusion Policy, $\pi_0$, and GR00T N1) across all robot embodiments for dexterous tasks involving tool manipulation, manipulation with deformable objects, and pick-and-place. Importantly, these tasks present significant simulation challenges due to their complex physical interactions with tools and deformable materials, making synthetic data generation infeasible with current approaches in the literature. Empricially, we observe a higher performance gain for GR00T N1 compared to DP and $\pi_0$; we hypothesize that having separate action and decoder parameters for the IDM actions help with the fact that neural trajectories have 0's as state.


\subsection{Unlocking Generalization}
\label{section:result_behavior}
To demonstrate how \ourmethod can unlock generalization in robot learning, we train our target video world model on 2,884 trajectories of the GR1 Humanoid performing diverse pick-and-place motions. Next, we prompt the model with (1) novel behaviors in seen environments and (2) seen and novel behaviors in novel environments, generating neural trajectories. The visualization of the evaluation configuration (how much randomization is done for the target object) is provided in Figure \ref{fig:eval_frames}. We use GR00T N1 as the base policy for this section.


\paragraph{Behavior Generalization}
We investigate whether our pipeline enables robots to learn entirely new behaviors \textit{solely} from neural trajectories without involving any human teleoperation. We define ``new behaviors" as novel action \textit{verbs} beyond adapting existing motions. Surprisingly, just given the initial frame and the language instruction, we observe that the video world model can generalize in generating videos of totally unseen behaviors (examples shown in Figure \ref{fig:sample_dreams}). We recommend referring to the website~\footnote{\url{https://research.nvidia.com/labs/gear/dreamgen}} for better visualizations. Leveraging this capability, we generate 50 neural trajectories for each of the 14 novel behavior tasks and train our downstream visuomotor robot policy only on the neural trajectories. As shown in Table \ref{tab:generalization}, we first show the result of GR00T N1 fine-tuned on the 2,885 pick-and-place trajectories, which also gets a somewhat non-trivial performance (11.8\%), due to some of the tasks giving partial points for picking up the object (e.g. for example, we give 0.5 success for picking up the bottle for the ``Pour Water" task). Nonetheless, we see a non-trivial performance gain when trained with neural trajectories (11.2\% $\rightarrow$ 43.2\%), showing that our pipeline enables learning totally new verbs.

\paragraph{Environment Generalization}

\begin{table}[t!]
\centering
\caption{\textbf{Success Rate (\%) Across New Behaviors (14 tasks) and Environments (13 tasks).}}
\resizebox{\textwidth}{!}{%
\begin{tabular}{l|cccccccccccccc|c}
\toprule
& \multicolumn{15}{c}{\textbf{Seen Environments, Novel Behaviors}} \\
\midrule
Model & \begin{tabular}[c]{@{}c@{}}Open\\Microwave\end{tabular} 
& \begin{tabular}[c]{@{}c@{}}Open\\Macbook\end{tabular} 
& \begin{tabular}[c]{@{}c@{}}Close \\Lunchbox\end{tabular} 
& \begin{tabular}[c]{@{}c@{}}Hit \\Tambourine\end{tabular} 
& \begin{tabular}[c]{@{}c@{}}Hit \\Keyboard\end{tabular} 
& \begin{tabular}[c]{@{}c@{}}Grab \\button\end{tabular} 
& \begin{tabular}[c]{@{}c@{}}Pour\\Water\end{tabular} 
& \begin{tabular}[c]{@{}c@{}}Water \\flowers\end{tabular} 
& \begin{tabular}[c]{@{}c@{}}Light\\Candle\end{tabular} 
& \begin{tabular}[c]{@{}c@{}}Use \\Vacuum\end{tabular} 
& \begin{tabular}[c]{@{}c@{}}Iron \\shirt\end{tabular} 
& \begin{tabular}[c]{@{}c@{}}Take Spoon\\Out\end{tabular} 
& \begin{tabular}[c]{@{}c@{}}Unroll \\mat\end{tabular} 
& \begin{tabular}[c]{@{}c@{}}Move \\Mouse\end{tabular} 
& Average \\
\midrule
GR00T N1 & 0 & 0 & 0 & 5 & 0 & 45 & 40 & 50 & 10 & 0 & 0 & 7 & 0 & 0 & 11.2 \\
\quad w/ \ourmethod & \textbf{23} & \textbf{45} & \textbf{10} & \textbf{15} & \textbf{90} & \textbf{75} & \textbf{55} & \textbf{95} & \textbf{15} & \textbf{55} & \textbf{20} & \textbf{17} & \textbf{55} & \textbf{35} & \textbf{43.2} \\
\midrule
Examples & \multicolumn{15}{c}{\raisebox{-0.5\height}{\addAugFig{1.7}{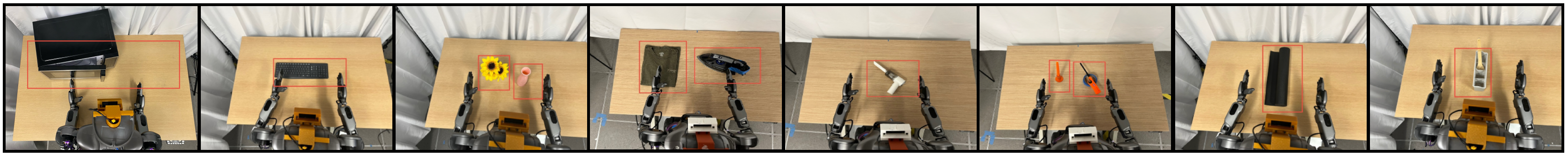}}} \\
\midrule
\end{tabular}
}

\resizebox{\textwidth}{!}{%
\begin{tabular}{L{2.84cm}|cccccc|ccccccc|c}
& \multicolumn{6}{c|}{\textbf{Novel Environments, Seen Behaviors}} & \multicolumn{7}{c|}{\textbf{Novel Environments, Novel Behaviors}} & \\
\midrule
Model & \begin{tabular}[c]{@{}c@{}}Pick up\\Tangerine\end{tabular} & \begin{tabular}[c]{@{}c@{}}Box \\sandwich\end{tabular} & \begin{tabular}[c]{@{}c@{}}Weigh the\\Orange\end{tabular} & \begin{tabular}[c]{@{}c@{}}Put cup\\in trash\end{tabular} & \begin{tabular}[c]{@{}c@{}}Put pear\\in basket\end{tabular} & \begin{tabular}[c]{@{}c@{}}Put sauce\\ on tray\end{tabular} & \begin{tabular}[c]{@{}c@{}}Water \\Flowers\end{tabular} & \begin{tabular}[c]{@{}c@{}}Lift \\Basket\end{tabular} & \begin{tabular}[c]{@{}c@{}}Swirl Around\\Spoon\end{tabular} & \begin{tabular}[c]{@{}c@{}}Use\\Whisk\end{tabular} & \begin{tabular}[c]{@{}c@{}}Close soup\\ container\end{tabular} & \begin{tabular}[c]{@{}c@{}}Uncover\\Pot\end{tabular} & \begin{tabular}[c]{@{}c@{}}Cover \\Pot\end{tabular} & Average \\
\midrule
GR00T N1 & 0 & 0 & 0 & 0 & 0 & 0 & 0 & 0 & 0 & 0 & 0 & 0 & 0 & 0.0 \\
\quad w/ \ourmethod & \textbf{30} & \textbf{10} & \textbf{20} & \textbf{45} & \textbf{35} & \textbf{45} & \textbf{15} & \textbf{55} & \textbf{15} & \textbf{25} & \textbf{55} & \textbf{30} & \textbf{35} & \textbf{28.5} \\
\midrule
Examples & \multicolumn{14}{c}{
\raisebox{-0.5\height}{\addAugFig{1.7}{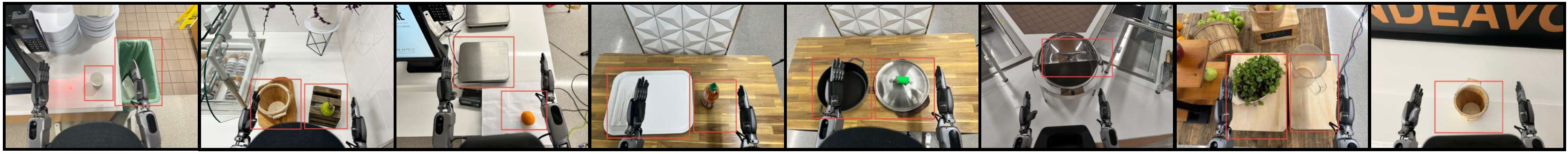}}
} \\
\bottomrule
\end{tabular}
}
\label{tab:generalization}
\end{table}

To our surprise, when prompted with initial frames of totally new environments, we observe that video world models can still generalize and generate very realistic robot videos, following the kinematics it learned during fine-tuning, while retaining the internet-video knowledge learned during pretraining. We follow the same proposed pipeline and train visuomotor robot policies \textit{solely} on neural trajectories, and observe that we can get non-trivial success rates on both seen behaviors (variants of pick-and-place) and unseen behaviors (e.g., watering flowers, closing containers, stirring whisk, etc.) as shown in Table \ref{tab:generalization}. Importantly, unlike previous work that showed environment generalization by scaling the total number of environments in the training data~\citep{intelligence2025pi}, our approach did not require any physical data collection beyond a single environment (i.e., lab setup)—we only capture initial frames, effectively implementing a \textit{zero-shot} transfer methodology. Lastly, the baseline model trained only on pick-and-place in a single environment shows 0\% Success Rate, since it does not have the ability to generalize beyond the environment it was trained in.

\section{\ourbench: A Video Generation Benchmark for Robotics}
\label{sec:wm_benchmark}
Motivated by recent work benchmarking the capabilities of video generative models as world models~\citep{kang2024far, bansal2024videophy, motamed2025generative, duan2025worldscore}, we introduce \ourbench, a systematic world modeling benchmark that aims to quantify the capacity of existing video generative models to adapt to a specific robot embodiment, internalizing the rigid body physics of the given robot, while generalizing to new objects, behaviors, and environments. We measure two key metrics: \textit{instruction following} and \textit{physics following}.

First, the \textit{instruction following} metric is used to assess whether the generated video strictly adheres to given instructions to generate a video of the robot \textit{completing} the specific task. The generated videos are fed into Qwen-VL-2.5~\citep{Qwen2.5-VL} with specific prompts to give a binary score (0 or 1) for quantifying the consistency between the video content and the task instructions, thereby ensuring that the actions and scenes in the video match the intended objectives. We provide the exact prompt we use for the evaluation in Appendix~\ref{sec:sr}. We also provide human evaluations in addition to the model-based evaluation, showing an average Pearson correlation of $\textgreater$ 90\%, ensuring that the model-based evaluation metric is aligned to human judgment in Appendix \ref{sec:human}.

Next, we quantify the \textit{physics alignment} to evaluate the physical plausibility of the generated videos, so that the videos are actually useful for downstream robot learning. For this purpose, we first employ the VideoCon-Physics~\citep{bansal2024videophy}, a VLM specifically trained to give scores for physics adherence of generated videos. Specifically, we get a 0 to 1 score from VideoCon-Physics. In practice, we find the model has not been trained on multiview videos (RoboCasa) and diverse robot environments, so we use a general VLM: Qwen-VL-2.5 to also score each video based on our instruction and then calculate the average score of these two scores for each video generation model on each dataset. We provide more details of VideoCon-Physics in Appendix \ref{sec:pa}. 


Using these two metrics, we benchmark 4 different video world models, Hunyuan~\citep{kong2024hunyuanvideo}, CogVideoX~\citep{yang2024cogvideox}, WAN 2.1~\citep{wang2025wan}, and Cosmos~\citep{agarwal2025cosmos}, on 2 different training and evaluation setups, one in simulation on the Franka Emika robot and one in real on the Fourier GR1 Humanoid. We also quantify the \textit{zero-shot} capability of the models, evaluated without adapting to the specific embodiment. Results and dataset statistics are shown in Table \ref{tab:wm_results}. In addition to these two metrics, we also replay the IDM actions in simulation to empirically see the quality of the IDM actions, where we have access to the digital twin of the Fourier GR1. See Section \ref{sec:intermediary} for more details.

\begin{table}[t!]
\centering
\caption{\textbf{\ourbench Statistics and Results}. IF represents Instruction Following, and PA represents Physics Alignment. GPT represents the evaluation from GPT4o, Qwen represents the evaluation from Qwen2.5VL, and Hu represents the human evaluation. -zero represents zero-shot inference and -sft represents fine-tuned variants. Best is \textbf{bolded} and second best is \underline{underlined}.}
\label{tab:combined}
\resizebox{\textwidth}{!}{%
\begin{tabular}{l|ccc|c|ccc|c|ccc|c|ccc|c}
\toprule
\multicolumn{17}{c}{\textbf{Dataset Statistics}} \\
\midrule
Dataset & \multicolumn{4}{c|}{RoboCasa} & \multicolumn{12}{c}{GR1} \\
\midrule
Train (\# trajs) & \multicolumn{4}{c|}{1200} & \multicolumn{12}{c}{100} \\
\midrule
Eval (\# frames) & \multicolumn{4}{c|}{48} & \multicolumn{4}{c|}{Object: 50} & \multicolumn{4}{c|}{Behavior: 47} & \multicolumn{4}{c}{Env: 30} \\
\toprule
\multicolumn{17}{c}{\textbf{Results}} \\
\midrule
& \multicolumn{3}{c|}{\textbf{IF}} & \textbf{PA} & \multicolumn{3}{c|}{\textbf{IF}} & \textbf{PA} & \multicolumn{3}{c|}{\textbf{IF}} & \textbf{PA} & \multicolumn{3}{c|}{\textbf{IF}} & \textbf{PA} \\
 & GPT & Qwen & Hu & & GPT & Qwen & Hu & & GPT & Qwen & Hu & & GPT & Qwen & Hu & \\
\midrule
Hunyuan-zero    & \underline{1.0} & 0.0   & - & 0.0 & 0.0 & 0.0 & - & 0.0 & 0.0 & \underline{2.1} & - & \underline{2.1} & 0.0 & 0.0 & - & 0.0 \\
CogVideoX-zero  & 0.0             & 0.0   & - & 0.0 & 0.0 & 0.0 & - & 0.0 & 0.0             & 0.0             & - & 0.0             & 0.0           & 0.0 & - & 0.0 \\
WAN2.1-zero     & 0.0             & 0.0   & - & 0.0 & 0.0 & \underline{2.0} & - & \underline{2.0} & 0.0             & \underline{2.1} & - & \underline{2.1} & 0.0           & \underline{6.7} & - & \underline{6.7} \\
Cosmos-zero     & \textbf{4.2}    & \textbf{22.9} & - & \textbf{22.9} & 0.0 & \textbf{32.0} & - & \textbf{32.0} & \textbf{6.4}    & \textbf{31.9} & - & \textbf{31.9} & \textbf{3.5} & \textbf{24.1} & - & \textbf{24.1} \\
\midrule
Hunyuan-sft     & 68.8   &  8.3  & 81.3 & 44.8 & 38.0 & 26.0 & 52.0 & 39.0 & 38.3 & 10.6 & 14.9 & 12.8 & 27.6 & 27.6 & 43.2 & 35.4 \\
CogVideoX-sft   & 72.9   & 10.4  & 79.2 & 44.8 & \underline{72.0} & 38.0 & 72.0 & 55.0 & 44.0 & 28.0 & 21.3 & \underline{24.7} & \underline{55.2} & \underline{41.4} & \underline{61.1} & 51.3 \\
WAN2.1-sft      & \underline{77.1} & \underline{18.8} & \underline{91.7} & \underline{55.3} & \underline{72.0} & \underline{58.0} & \underline{80.0} & \underline{69.0} & \textbf{72.3} & \underline{55.3} & \textbf{74.5} & \textbf{64.9} & 48.3 & \textbf{65.5} & \textbf{67.4} & \textbf{66.5} \\
Cosmos-sft      & \textbf{79.2}    & \textbf{29.2} & \textbf{93.8} & \textbf{61.5} & \textbf{90.0} & \textbf{62.0} & \textbf{84.0} & \textbf{73.0} & \underline{59.6} & \textbf{61.7} & \underline{68.1} & \textbf{64.9} & \textbf{69.0} & \textbf{65.5} & 53.3 & \underline{59.4} \\
\bottomrule
\end{tabular}
\label{tab:wm_results}
}
\vspace{-5mm}
\end{table}




\begin{wrapfigure}{r}{0.4\textwidth}
    \vspace{-4mm}
    \centering
    \includegraphics[width=\linewidth]{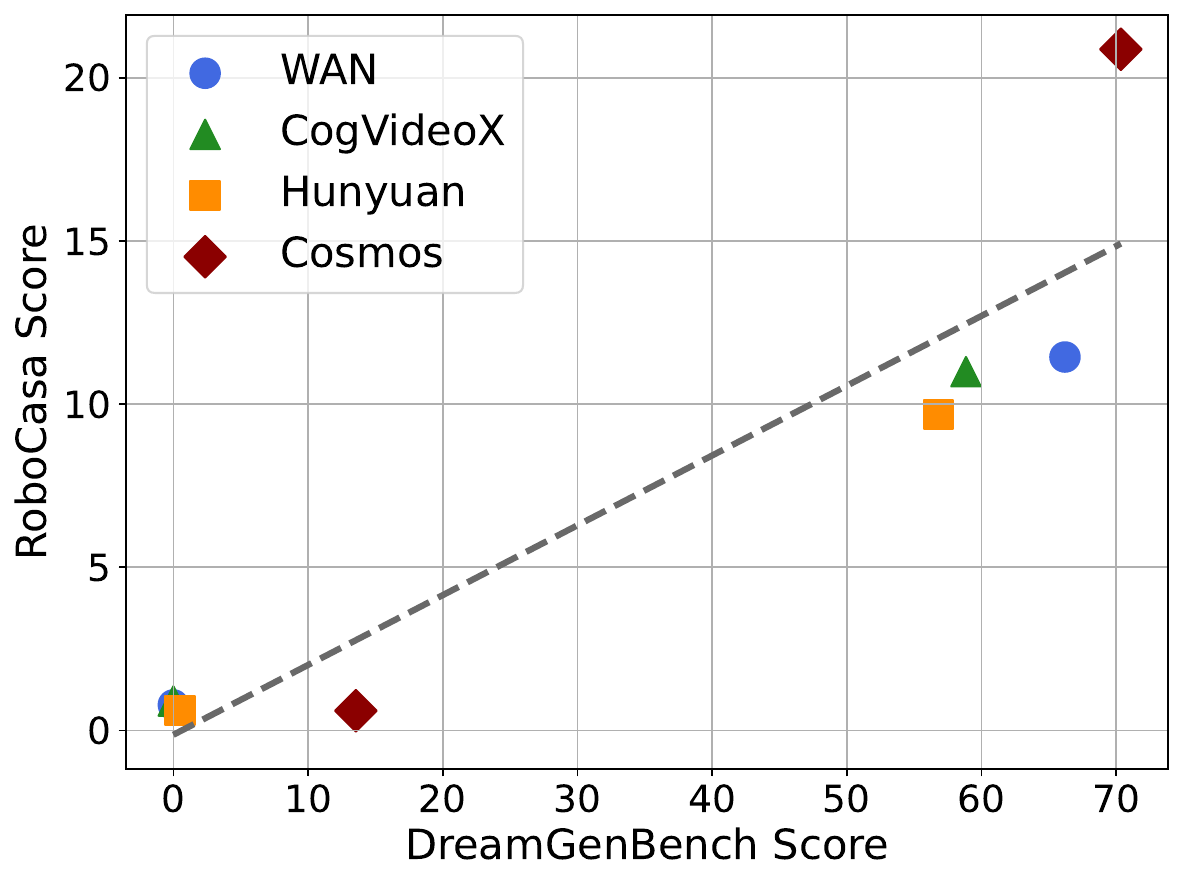}
\caption{Performance correlation between \ourbench and RoboCasa.}
\label{fig:scatter_plot}
\end{wrapfigure} 

\paragraph{\ourbench shows positive correlation to downstream robot policy performance.}
To measure whether \ourbench could be a proxy evaluation for the performance of the downstream robot policy, we measure the performance of the RoboCasa benchmark by \textit{only} training on neural trajectories generated from the different video world models. A positive correlation between \ourbench and RoboCasa would indicate that building a better world model that can follow language instruction and model world physics leads to better performance on the downstream robot manipulation tasks. We compare all the models in Table \ref{tab:wm_results} with 7K neural trajectories per model. For \ourbench score, we use the average of IF (GPT) and PA from Table \ref{tab:wm_results}. The results are illustrated in Figure \ref{fig:scatter_plot}. As shown, the correlation between \ourbench and RoboCasa shows a positive correlation, indicating that building a stronger video world model could lead to larger performance enhancement.

\section{Related Work}
\paragraph{Synthetic Data Generation in Robotics.}
Real-world robot data collection through human teleoperation requires large amounts of time and considerable human cost. As an alternative, collecting synthetic data in simulation can be more efficient and automated with minimal human effort ~\citep{mandlekar2023mimicgen, james2020rlbench, dalal2023imitating, gu2023maniskill2, ha2023scaling, robocasa2024, jiang2024dexmimicen, wang2023robogen, su2019dynamic, garrett2024skillmimicgen, yang2025physics}. However, using these trajectories can be challenging due to the following factors: (1) the sim-to-real gap, (2) difficulty in simulating objects such as liquid and articulated objects, and (3) being bounded by either Task and Motion Planning (TAMP) based systems or the interpolation of human teleoperation data.
Another direction is to use neural generative models to augment existing sets of robot demonstrations~\citep{mandi2022cacti, yu2023scaling, chen2023genaug, chen2024rovi}, using in-painting, image diffusion models, or even video2video models~\citep{alhaija2025cosmos}. However, the diversity of the generated data is limited, especially in terms of robot motions, and the augmented data is only used to increase visual robustness to distribution shifts. 

\paragraph{Video World Modeling for Robotics.} Video generative models can be used to generate synthetic robot trajectories and extract executable actions during test-time via inverse-dynamics models (IDM)~\citep{du2023learning, zhou2024robodreamer}, optical flow as dense correspondence~\citep{ko2024learning}, or trajectories as high-level plans~\citep{yang2024learning, du2024video}. Another work generates human videos along with 3D tracking during test-time~\citep{liang2024dreamitate}, or human videos for novel scenes and motions~\citep{bharadhwaj2024gen2act}, and trains a policy with a point tracking objective. A concurrent work explores adapting text-to-video models for task generalization~\citep{luo2025solving} by generating synthetic trajectories and extracting executable actions via an IDM or using it to extract rewards to guide a reinforcement learning policy. However, the scope of the work is bounded by simulation tasks. 
Some recent work aims to either train a robot policy initialized from a video generative model~\citep{wu2023unleashing, cheang2024gr} or perform policy training, inverse dynamics, and forward dynamics together, enabling co-training with both robot and video data~\citep{guo2024prediction, li2025unified, zhu2025unified, zhao2025cot}. Our approach deliberately separates these components to fully make use of the state-of-the-art video generative models, which is currently not feasible to run in adjacent with a robot policy real time to ensure the strongest generalization capabilites.

\paragraph{Learning Robot Policies from Videos}
Videos provide abundant information for training robots, yet most do not come with labeled actions~\citep{mccarthy2024generalistrobotlearninginternet}. To enhance visual representations, prior work has used pretraining of vision encoders on egocentric videos of human activity~\citep{grauman2022ego4d}, which has proven beneficial in downstream tasks~\citep{nair2022r3m, dasari2023unbiased}. Several approaches extract various forms of information from human-centric videos, including human-object interactions~\citep{zeng2024learning}, object affordances~\citep{bahl2023affordances, kannan2023deft, srirama2024hrp, shaw2023videodex}, and visual trajectories~\citep{wen2023any, bharadhwaj2024track2act}. Other lines of research focus on translating human motions into robotic behaviors, employing hand pose estimators~\citep{wang2023mimicplay, zhu2024vision, shaw2023videodex, bharadhwaj2023zero, ye2023learning, qin2022dexmv} or motion capture systems~\citep{yang2024equibot}. Another line of work extracts \textit{latent} actions to train downstream robot policies from visual deltas between the current and future frames~\citep{ye2025latent, bruce2024geniegenerativeinteractiveenvironments, chen2025motolatentmotiontoken, schmidt2024learning, ren2025videoworldexploringknowledgelearning, bu2025agibot, bu2025univla, gao2025adaworld}. In this work, we use \textit{synthetic} videos generated by a world model as the source instead of human videos, and explore using latent actions by co-training latent actions with real-world actions. 
\section{Conclusion}

We introduce a novel pipeline for robot learning that taps into the power of SOTA video generative models. By generating synthetic videos and extracting pseudo-actions, we enable training visuomotor policies without relying solely on manual demonstrations. This approach not only augments existing tasks but also unlocks the ability to learn entirely new behaviors in unseen environments. \ourmethod serves as a solid stepping stone towards unleashing the full potential of world models in robotics.

\section{Limitation}
Our approach is complementary to existing methods that learn from videos, although we do not directly benchmark against them. Many of these works focus on learning from human demonstration videos. Since \ourmethod helps bridge the human-robot domain gap, we believe it can serve as a useful foundation for improving such methods and enabling broader generalization. Our tasks are relatively simple and cover a limited portion of the robot's full kinematic capabilities. Supporting more complex, dexterous behaviors that require richer control remains an important direction for future work. Increasing the diversity of training behaviors, along with broader video-language pairings, may allow the video world model to take on more of the representational burden and improve generalization to challenging tasks.

\ourmethod currently requires significant compute. For instance, generating the 240k-sample RoboCasa dataset took 54 hours on 1500 NVIDIA L40 GPUs. While feasible in a large-scale research setting, reducing computational cost without sacrificing the strength of video priors remains an important challenge. The method also relies on manually providing initial frames, which introduces operational overhead. Developing automated ways to generate or select initial frames is a promising future direction.

Finally, the automatic evaluator used in \ourbench is based on lightweight open-source models to keep the benchmark accessible. These models can occasionally hallucinate, especially when evaluating physical realism in videos, which remains a difficult and evolving problem. We acknowledge this limitation and leave improvements in evaluation to future work.




\bibliography{example}  

\begin{thebibliography}{78}
\providecommand{\natexlab}[1]{#1}
\providecommand{\url}[1]{\texttt{#1}}
\expandafter\ifx\csname urlstyle\endcsname\relax
  \providecommand{\doi}[1]{doi: #1}\else
  \providecommand{\doi}{doi: \begingroup \urlstyle{rm}\Url}\fi

\bibitem[Brohan et~al.(2023)Brohan, Brown, Carbajal, Chebotar, Chen, Choromanski, Ding, Driess, Dubey, Finn, Florence, Fu, Arenas, Gopalakrishnan, Han, Hausman, Herzog, Hsu, Ichter, Irpan, Joshi, Julian, Kalashnikov, Kuang, Leal, Lee, Lee, Levine, Lu, Michalewski, Mordatch, Pertsch, Rao, Reymann, Ryoo, Salazar, Sanketi, Sermanet, Singh, Singh, Soricut, Tran, Vanhoucke, Vuong, Wahid, Welker, Wohlhart, Wu, Xia, Xiao, Xu, Xu, Yu, and Zitkovich]{rt22023arxiv}
A.~Brohan, N.~Brown, J.~Carbajal, Y.~Chebotar, X.~Chen, K.~Choromanski, T.~Ding, D.~Driess, A.~Dubey, C.~Finn, P.~Florence, C.~Fu, M.~G. Arenas, K.~Gopalakrishnan, K.~Han, K.~Hausman, A.~Herzog, J.~Hsu, B.~Ichter, A.~Irpan, N.~Joshi, R.~Julian, D.~Kalashnikov, Y.~Kuang, I.~Leal, L.~Lee, T.-W.~E. Lee, S.~Levine, Y.~Lu, H.~Michalewski, I.~Mordatch, K.~Pertsch, K.~Rao, K.~Reymann, M.~Ryoo, G.~Salazar, P.~Sanketi, P.~Sermanet, J.~Singh, A.~Singh, R.~Soricut, H.~Tran, V.~Vanhoucke, Q.~Vuong, A.~Wahid, S.~Welker, P.~Wohlhart, J.~Wu, F.~Xia, T.~Xiao, P.~Xu, S.~Xu, T.~Yu, and B.~Zitkovich.
\newblock Rt-2: Vision-language-action models transfer web knowledge to robotic control.
\newblock In \emph{arXiv preprint arXiv:2307.15818}, 2023.

\bibitem[Black et~al.()Black, Brown, Driess, Esmail, Equi, Finn, Fusai, Groom, Hausman, Ichter, et~al.]{black2410pi0}
K.~Black, N.~Brown, D.~Driess, A.~Esmail, M.~Equi, C.~Finn, N.~Fusai, L.~Groom, K.~Hausman, B.~Ichter, et~al.
\newblock $\pi$0: A vision-language-action flow model for general robot control, 2024.
\newblock \emph{URL https://arxiv. org/abs/2410.24164}.

\bibitem[Team et~al.(2025)Team, Abeyruwan, Ainslie, Alayrac, Arenas, Armstrong, Balakrishna, Baruch, Bauza, Blokzijl, et~al.]{team2025gemini}
G.~R. Team, S.~Abeyruwan, J.~Ainslie, J.-B. Alayrac, M.~G. Arenas, T.~Armstrong, A.~Balakrishna, R.~Baruch, M.~Bauza, M.~Blokzijl, et~al.
\newblock Gemini robotics: Bringing ai into the physical world.
\newblock \emph{arXiv preprint arXiv:2503.20020}, 2025.

\bibitem[Bu et~al.(2025)Bu, Cai, Chen, Cui, Ding, Feng, Gao, He, Huang, Jiang, et~al.]{bu2025agibot}
Q.~Bu, J.~Cai, L.~Chen, X.~Cui, Y.~Ding, S.~Feng, S.~Gao, X.~He, X.~Huang, S.~Jiang, et~al.
\newblock Agibot world colosseo: A large-scale manipulation platform for scalable and intelligent embodied systems.
\newblock \emph{arXiv preprint arXiv:2503.06669}, 2025.

\bibitem[Bjorck et~al.(2025)Bjorck, Casta{\~n}eda, Cherniadev, Da, Ding, Fan, Fang, Fox, Hu, Huang, et~al.]{bjorck2025gr00t}
J.~Bjorck, F.~Casta{\~n}eda, N.~Cherniadev, X.~Da, R.~Ding, L.~Fan, Y.~Fang, D.~Fox, F.~Hu, S.~Huang, et~al.
\newblock Gr00t n1: An open foundation model for generalist humanoid robots.
\newblock \emph{arXiv preprint arXiv:2503.14734}, 2025.

\bibitem[Intelligence et~al.(2025)Intelligence, Black, Brown, Darpinian, Dhabalia, Driess, Esmail, Equi, Finn, Fusai, et~al.]{intelligence2025pi}
P.~Intelligence, K.~Black, N.~Brown, J.~Darpinian, K.~Dhabalia, D.~Driess, A.~Esmail, M.~Equi, C.~Finn, N.~Fusai, et~al.
\newblock $\pi_{0.5}$: a vision-language-action model with open-world generalization.
\newblock \emph{arXiv preprint arXiv:2504.16054}, 2025.

\bibitem[Agarwal et~al.(2025)Agarwal, Ali, Bala, Balaji, Barker, Cai, Chattopadhyay, Chen, Cui, Ding, et~al.]{agarwal2025cosmos}
N.~Agarwal, A.~Ali, M.~Bala, Y.~Balaji, E.~Barker, T.~Cai, P.~Chattopadhyay, Y.~Chen, Y.~Cui, Y.~Ding, et~al.
\newblock Cosmos world foundation model platform for physical ai.
\newblock \emph{arXiv preprint arXiv:2501.03575}, 2025.

\bibitem[Yang et~al.(2024)Yang, Teng, Zheng, Ding, Huang, Xu, Yang, Hong, Zhang, Feng, et~al.]{yang2024cogvideox}
Z.~Yang, J.~Teng, W.~Zheng, M.~Ding, S.~Huang, J.~Xu, Y.~Yang, W.~Hong, X.~Zhang, G.~Feng, et~al.
\newblock Cogvideox: Text-to-video diffusion models with an expert transformer.
\newblock \emph{arXiv preprint arXiv:2408.06072}, 2024.

\bibitem[Wang et~al.(2025)Wang, Ai, Wen, Mao, Xie, Chen, Yu, Zhao, Yang, Zeng, et~al.]{wang2025wan}
A.~Wang, B.~Ai, B.~Wen, C.~Mao, C.-W. Xie, D.~Chen, F.~Yu, H.~Zhao, J.~Yang, J.~Zeng, et~al.
\newblock Wan: Open and advanced large-scale video generative models.
\newblock \emph{arXiv preprint arXiv:2503.20314}, 2025.

\bibitem[Kong et~al.(2024)Kong, Tian, Zhang, Min, Dai, Zhou, Xiong, Li, Wu, Zhang, et~al.]{kong2024hunyuanvideo}
W.~Kong, Q.~Tian, Z.~Zhang, R.~Min, Z.~Dai, J.~Zhou, J.~Xiong, X.~Li, B.~Wu, J.~Zhang, et~al.
\newblock Hunyuanvideo: A systematic framework for large video generative models.
\newblock \emph{arXiv preprint arXiv:2412.03603}, 2024.

\bibitem[Lin et~al.(2024)Lin, Liu, Chen, Lu, Hu, Fu, Allardice, Lai, Song, Zhang, et~al.]{lin2024stiv}
Z.~Lin, W.~Liu, C.~Chen, J.~Lu, W.~Hu, T.-J. Fu, J.~Allardice, Z.~Lai, L.~Song, B.~Zhang, et~al.
\newblock Stiv: Scalable text and image conditioned video generation.
\newblock \emph{arXiv preprint arXiv:2412.07730}, 2024.

\bibitem[Xiang et~al.(2024)Xiang, Liu, Gu, Gao, Ning, Zha, Feng, Tao, Hao, Shi, et~al.]{xiang2024pandora}
J.~Xiang, G.~Liu, Y.~Gu, Q.~Gao, Y.~Ning, Y.~Zha, Z.~Feng, T.~Tao, S.~Hao, Y.~Shi, et~al.
\newblock Pandora: Towards general world model with natural language actions and video states.
\newblock \emph{arXiv preprint arXiv:2406.09455}, 2024.

\bibitem[Ye et~al.(2025)Ye, Jang, Jeon, Joo, Yang, Peng, Mandlekar, Tan, Chao, Lin, Liden, Lee, Gao, Zettlemoyer, Fox, and Seo]{ye2025latent}
S.~Ye, J.~Jang, B.~Jeon, S.~J. Joo, J.~Yang, B.~Peng, A.~Mandlekar, R.~Tan, Y.-W. Chao, B.~Y. Lin, L.~Liden, K.~Lee, J.~Gao, L.~Zettlemoyer, D.~Fox, and M.~Seo.
\newblock Latent action pretraining from videos.
\newblock In \emph{The Thirteenth International Conference on Learning Representations}, 2025.
\newblock URL \url{https://openreview.net/forum?id=VYOe2eBQeh}.

\bibitem[Baker et~al.(2022)Baker, Akkaya, Zhokov, Huizinga, Tang, Ecoffet, Houghton, Sampedro, and Clune]{baker2022video}
B.~Baker, I.~Akkaya, P.~Zhokov, J.~Huizinga, J.~Tang, A.~Ecoffet, B.~Houghton, R.~Sampedro, and J.~Clune.
\newblock Video pretraining (vpt): Learning to act by watching unlabeled online videos.
\newblock \emph{Advances in Neural Information Processing Systems}, 35:\penalty0 24639--24654, 2022.

\bibitem[Du et~al.(2023)Du, Yang, Dai, Dai, Nachum, Tenenbaum, Schuurmans, and Abbeel]{du2023learning}
Y.~Du, S.~Yang, B.~Dai, H.~Dai, O.~Nachum, J.~Tenenbaum, D.~Schuurmans, and P.~Abbeel.
\newblock Learning universal policies via text-guided video generation.
\newblock \emph{Advances in neural information processing systems}, 36:\penalty0 9156--9172, 2023.

\bibitem[Zhou et~al.(2024)Zhou, Du, Chen, Li, Yeung, and Gan]{zhou2024robodreamer}
S.~Zhou, Y.~Du, J.~Chen, Y.~Li, D.-Y. Yeung, and C.~Gan.
\newblock Robodreamer: Learning compositional world models for robot imagination.
\newblock \emph{arXiv preprint arXiv:2404.12377}, 2024.

\bibitem[Ko et~al.(2024)Ko, Mao, Du, Sun, and Tenenbaum]{ko2024learning}
P.-C. Ko, J.~Mao, Y.~Du, S.-H. Sun, and J.~B. Tenenbaum.
\newblock Learning to act from actionless videos through dense correspondences.
\newblock In \emph{The Twelfth International Conference on Learning Representations}, 2024.
\newblock URL \url{https://openreview.net/forum?id=Mhb5fpA1T0}.

\bibitem[Yang et~al.(2024)Yang, Du, Ghasemipour, Tompson, Kaelbling, Schuurmans, and Abbeel]{yang2024learning}
S.~Yang, Y.~Du, S.~K.~S. Ghasemipour, J.~Tompson, L.~P. Kaelbling, D.~Schuurmans, and P.~Abbeel.
\newblock Learning interactive real-world simulators.
\newblock In \emph{The Twelfth International Conference on Learning Representations}, 2024.
\newblock URL \url{https://openreview.net/forum?id=sFyTZEqmUY}.

\bibitem[Du et~al.(2024)Du, Yang, Florence, Xia, Wahid, brian ichter, Sermanet, Yu, Abbeel, Tenenbaum, Kaelbling, Zeng, and Tompson]{du2024video}
Y.~Du, S.~Yang, P.~Florence, F.~Xia, A.~Wahid, brian ichter, P.~Sermanet, T.~Yu, P.~Abbeel, J.~B. Tenenbaum, L.~P. Kaelbling, A.~Zeng, and J.~Tompson.
\newblock Video language planning.
\newblock In \emph{The Twelfth International Conference on Learning Representations}, 2024.
\newblock URL \url{https://openreview.net/forum?id=9pKtcJcMP3}.

\bibitem[Nasiriany et~al.(2024)Nasiriany, Maddukuri, Zhang, Parikh, Lo, Joshi, Mandlekar, and Zhu]{robocasa2024}
S.~Nasiriany, A.~Maddukuri, L.~Zhang, A.~Parikh, A.~Lo, A.~Joshi, A.~Mandlekar, and Y.~Zhu.
\newblock Robocasa: Large-scale simulation of everyday tasks for generalist robots.
\newblock In \emph{Robotics: Science and Systems (RSS)}, 2024.

\bibitem[Hu et~al.(2022)Hu, Shen, Wallis, Allen-Zhu, Li, Wang, Wang, Chen, et~al.]{hu2022lora}
E.~J. Hu, Y.~Shen, P.~Wallis, Z.~Allen-Zhu, Y.~Li, S.~Wang, L.~Wang, W.~Chen, et~al.
\newblock Lora: Low-rank adaptation of large language models.
\newblock \emph{ICLR}, 1\penalty0 (2):\penalty0 3, 2022.

\bibitem[Khazatsky et~al.(2024)Khazatsky, Pertsch, Nair, Balakrishna, Dasari, Karamcheti, Nasiriany, Srirama, Chen, Ellis, et~al.]{khazatsky2024droid}
A.~Khazatsky, K.~Pertsch, S.~Nair, A.~Balakrishna, S.~Dasari, S.~Karamcheti, S.~Nasiriany, M.~K. Srirama, L.~Y. Chen, K.~Ellis, et~al.
\newblock Droid: A large-scale in-the-wild robot manipulation dataset.
\newblock \emph{arXiv preprint arXiv:2403.12945}, 2024.

\bibitem[Baker et~al.(2022)Baker, Akkaya, Zhokhov, Huizinga, Tang, Ecoffet, Houghton, Sampedro, and Clune]{baker2022videopretrainingvptlearning}
B.~Baker, I.~Akkaya, P.~Zhokhov, J.~Huizinga, J.~Tang, A.~Ecoffet, B.~Houghton, R.~Sampedro, and J.~Clune.
\newblock Video pretraining (vpt): Learning to act by watching unlabeled online videos, 2022.
\newblock URL \url{https://arxiv.org/abs/2206.11795}.

\bibitem[Chi et~al.(2023)Chi, Xu, Feng, Cousineau, Du, Burchfiel, Tedrake, and Song]{chi2023diffusion}
C.~Chi, Z.~Xu, S.~Feng, E.~Cousineau, Y.~Du, B.~Burchfiel, R.~Tedrake, and S.~Song.
\newblock Diffusion policy: Visuomotor policy learning via action diffusion.
\newblock \emph{The International Journal of Robotics Research}, page 02783649241273668, 2023.

\bibitem[Kang et~al.(2024)Kang, Yue, Lu, Lin, Zhao, Wang, Huang, and Feng]{kang2024far}
B.~Kang, Y.~Yue, R.~Lu, Z.~Lin, Y.~Zhao, K.~Wang, G.~Huang, and J.~Feng.
\newblock How far is video generation from world model: A physical law perspective.
\newblock \emph{arXiv preprint arXiv:2411.02385}, 2024.

\bibitem[Bansal et~al.(2024)Bansal, Lin, Xie, Zong, Yarom, Bitton, Jiang, Sun, Chang, and Grover]{bansal2024videophy}
H.~Bansal, Z.~Lin, T.~Xie, Z.~Zong, M.~Yarom, Y.~Bitton, C.~Jiang, Y.~Sun, K.-W. Chang, and A.~Grover.
\newblock Videophy: Evaluating physical commonsense for video generation.
\newblock \emph{arXiv preprint arXiv:2406.03520}, 2024.

\bibitem[Motamed et~al.(2025)Motamed, Culp, Swersky, Jaini, and Geirhos]{motamed2025generative}
S.~Motamed, L.~Culp, K.~Swersky, P.~Jaini, and R.~Geirhos.
\newblock Do generative video models learn physical principles from watching videos?
\newblock \emph{arXiv preprint arXiv:2501.09038}, 2025.

\bibitem[Duan et~al.(2025)Duan, Yu, Chen, Fei-Fei, and Wu]{duan2025worldscore}
H.~Duan, H.-X. Yu, S.~Chen, L.~Fei-Fei, and J.~Wu.
\newblock Worldscore: A unified evaluation benchmark for world generation.
\newblock \emph{arXiv preprint arXiv:2504.00983}, 2025.

\bibitem[Bai et~al.(2025)Bai, Chen, Liu, Wang, Ge, Song, Dang, Wang, Wang, Tang, Zhong, Zhu, Yang, Li, Wan, Wang, Ding, Fu, Xu, Ye, Zhang, Xie, Cheng, Zhang, Yang, Xu, and Lin]{Qwen2.5-VL}
S.~Bai, K.~Chen, X.~Liu, J.~Wang, W.~Ge, S.~Song, K.~Dang, P.~Wang, S.~Wang, J.~Tang, H.~Zhong, Y.~Zhu, M.~Yang, Z.~Li, J.~Wan, P.~Wang, W.~Ding, Z.~Fu, Y.~Xu, J.~Ye, X.~Zhang, T.~Xie, Z.~Cheng, H.~Zhang, Z.~Yang, H.~Xu, and J.~Lin.
\newblock Qwen2.5-vl technical report.
\newblock \emph{arXiv preprint arXiv:2502.13923}, 2025.

\bibitem[Mandlekar et~al.(2023)Mandlekar, Nasiriany, Wen, Akinola, Narang, Fan, Zhu, and Fox]{mandlekar2023mimicgen}
A.~Mandlekar, S.~Nasiriany, B.~Wen, I.~Akinola, Y.~Narang, L.~Fan, Y.~Zhu, and D.~Fox.
\newblock Mimicgen: A data generation system for scalable robot learning using human demonstrations.
\newblock In \emph{Conference on Robot Learning}, 2023.

\bibitem[James et~al.(2020)James, Ma, Arrojo, and Davison]{james2020rlbench}
S.~James, Z.~Ma, D.~R. Arrojo, and A.~J. Davison.
\newblock Rlbench: The robot learning benchmark \& learning environment.
\newblock \emph{IEEE Robotics and Automation Letters}, 5\penalty0 (2):\penalty0 3019--3026, 2020.

\bibitem[Dalal et~al.(2023)Dalal, Mandlekar, Garrett, Handa, Salakhutdinov, and Fox]{dalal2023imitating}
M.~Dalal, A.~Mandlekar, C.~R. Garrett, A.~Handa, R.~Salakhutdinov, and D.~Fox.
\newblock Imitating task and motion planning with visuomotor transformers.
\newblock In \emph{Conference on Robot Learning}, pages 2565--2593. PMLR, 2023.

\bibitem[Gu et~al.(2023)Gu, Xiang, Li, Ling, Liu, Mu, Tang, Tao, Wei, Yao, et~al.]{gu2023maniskill2}
J.~Gu, F.~Xiang, X.~Li, Z.~Ling, X.~Liu, T.~Mu, Y.~Tang, S.~Tao, X.~Wei, Y.~Yao, et~al.
\newblock Maniskill2: A unified benchmark for generalizable manipulation skills.
\newblock In \emph{The Eleventh International Conference on Learning Representations}, 2023.

\bibitem[Ha et~al.(2023)Ha, Florence, and Song]{ha2023scaling}
H.~Ha, P.~Florence, and S.~Song.
\newblock Scaling up and distilling down: Language-guided robot skill acquisition.
\newblock In \emph{Conference on Robot Learning}, pages 3766--3777. PMLR, 2023.

\bibitem[Jiang et~al.(2024)Jiang, Xie, Lin, Xu, Wan, Mandlekar, Fan, and Zhu]{jiang2024dexmimicen}
Z.~Jiang, Y.~Xie, K.~Lin, Z.~Xu, W.~Wan, A.~Mandlekar, L.~Fan, and Y.~Zhu.
\newblock Dexmimicgen: Automated data generation for bimanual dexterous manipulation via imitation learning.
\newblock 2024.

\bibitem[Wang et~al.(2024)Wang, Xian, Chen, Wang, Wang, Fragkiadaki, Erickson, Held, and Gan]{wang2023robogen}
Y.~Wang, Z.~Xian, F.~Chen, T.-H. Wang, Y.~Wang, K.~Fragkiadaki, Z.~Erickson, D.~Held, and C.~Gan.
\newblock Robogen: Towards unleashing infinite data for automated robot learning via generative simulation.
\newblock In \emph{International Conference on Machine Learning}, 2024.

\bibitem[Su et~al.(2019)Su, Zhou, Wu, Su, Liang, Liu, Zheng, Wang, Yan, and Hu]{su2019dynamic}
Y.~Su, S.~Zhou, Y.~Wu, T.~Su, D.~Liang, J.~Liu, D.~Zheng, Y.~Wang, J.~Yan, and X.~Hu.
\newblock Dynamic multi-path neural network.
\newblock \emph{arXiv preprint arXiv:1902.10949}, 2019.

\bibitem[Garrett et~al.(2024)Garrett, Mandlekar, Wen, and Fox]{garrett2024skillmimicgen}
C.~Garrett, A.~Mandlekar, B.~Wen, and D.~Fox.
\newblock Skillmimicgen: Automated demonstration generation for efficient skill learning and deployment.
\newblock \emph{arXiv preprint arXiv:2410.18907}, 2024.

\bibitem[Yang et~al.(2025)Yang, Suh, Zhao, Graesdal, Kelestemur, Wang, Pang, and Tedrake]{yang2025physics}
L.~Yang, H.~Suh, T.~Zhao, B.~P. Graesdal, T.~Kelestemur, J.~Wang, T.~Pang, and R.~Tedrake.
\newblock Physics-driven data generation for contact-rich manipulation via trajectory optimization.
\newblock \emph{arXiv preprint arXiv:2502.20382}, 2025.

\bibitem[Mandi et~al.(2022)Mandi, Bharadhwaj, Moens, Song, Rajeswaran, and Kumar]{mandi2022cacti}
Z.~Mandi, H.~Bharadhwaj, V.~Moens, S.~Song, A.~Rajeswaran, and V.~Kumar.
\newblock Cacti: A framework for scalable multi-task multi-scene visual imitation learning.
\newblock \emph{arXiv preprint arXiv:2212.05711}, 2022.

\bibitem[Yu et~al.(2023)Yu, Xiao, Stone, Tompson, Brohan, Wang, Singh, Tan, Peralta, Ichter, et~al.]{yu2023scaling}
T.~Yu, T.~Xiao, A.~Stone, J.~Tompson, A.~Brohan, S.~Wang, J.~Singh, C.~Tan, J.~Peralta, B.~Ichter, et~al.
\newblock Scaling robot learning with semantically imagined experience.
\newblock \emph{arXiv preprint arXiv:2302.11550}, 2023.

\bibitem[Chen et~al.(2023)Chen, Kiami, Gupta, and Kumar]{chen2023genaug}
Z.~Chen, S.~Kiami, A.~Gupta, and V.~Kumar.
\newblock Genaug: Retargeting behaviors to unseen situations via generative augmentation.
\newblock \emph{arXiv preprint arXiv:2302.06671}, 2023.

\bibitem[Chen et~al.(2024)Chen, Xu, Dharmarajan, Irshad, Cheng, Keutzer, Tomizuka, Vuong, and Goldberg]{chen2024rovi}
L.~Y. Chen, C.~Xu, K.~Dharmarajan, M.~Z. Irshad, R.~Cheng, K.~Keutzer, M.~Tomizuka, Q.~Vuong, and K.~Goldberg.
\newblock Rovi-aug: Robot and viewpoint augmentation for cross-embodiment robot learning.
\newblock \emph{arXiv preprint arXiv:2409.03403}, 2024.

\bibitem[Alhaija et~al.(2025)Alhaija, Alvarez, Bala, Cai, Cao, Cha, Chen, Chen, Ferroni, Fidler, et~al.]{alhaija2025cosmos}
H.~A. Alhaija, J.~Alvarez, M.~Bala, T.~Cai, T.~Cao, L.~Cha, J.~Chen, M.~Chen, F.~Ferroni, S.~Fidler, et~al.
\newblock Cosmos-transfer1: Conditional world generation with adaptive multimodal control.
\newblock \emph{arXiv preprint arXiv:2503.14492}, 2025.

\bibitem[Liang et~al.(2024)Liang, Liu, Ozguroglu, Sudhakar, Dave, Tokmakov, Song, and Vondrick]{liang2024dreamitate}
J.~Liang, R.~Liu, E.~Ozguroglu, S.~Sudhakar, A.~Dave, P.~Tokmakov, S.~Song, and C.~Vondrick.
\newblock Dreamitate: Real-world visuomotor policy learning via video generation.
\newblock In \emph{8th Annual Conference on Robot Learning}, 2024.
\newblock URL \url{https://openreview.net/forum?id=InT87E5sr4}.

\bibitem[Bharadhwaj et~al.(2024)Bharadhwaj, Dwibedi, Gupta, Tulsiani, Doersch, Xiao, Shah, Xia, Sadigh, and Kirmani]{bharadhwaj2024gen2act}
H.~Bharadhwaj, D.~Dwibedi, A.~Gupta, S.~Tulsiani, C.~Doersch, T.~Xiao, D.~Shah, F.~Xia, D.~Sadigh, and S.~Kirmani.
\newblock Gen2act: Human video generation in novel scenarios enables generalizable robot manipulation.
\newblock \emph{arXiv preprint arXiv:2409.16283}, 2024.

\bibitem[Luo et~al.(2025)Luo, Zeng, Du, and Sun]{luo2025solving}
C.~Luo, Z.~Zeng, Y.~Du, and C.~Sun.
\newblock Solving new tasks by adapting internet video knowledge.
\newblock In \emph{The Thirteenth International Conference on Learning Representations}, 2025.

\bibitem[Wu et~al.(2023)Wu, Jing, Cheang, Chen, Xu, Li, Liu, Li, and Kong]{wu2023unleashing}
H.~Wu, Y.~Jing, C.~Cheang, G.~Chen, J.~Xu, X.~Li, M.~Liu, H.~Li, and T.~Kong.
\newblock Unleashing large-scale video generative pre-training for visual robot manipulation.
\newblock \emph{arXiv preprint arXiv:2312.13139}, 2023.

\bibitem[Cheang et~al.(2024)Cheang, Chen, Jing, Kong, Li, Li, Liu, Wu, Xu, Yang, et~al.]{cheang2024gr}
C.-L. Cheang, G.~Chen, Y.~Jing, T.~Kong, H.~Li, Y.~Li, Y.~Liu, H.~Wu, J.~Xu, Y.~Yang, et~al.
\newblock Gr-2: A generative video-language-action model with web-scale knowledge for robot manipulation.
\newblock \emph{arXiv preprint arXiv:2410.06158}, 2024.

\bibitem[Guo et~al.(2024)Guo, Hu, Zhang, Wang, Chen, Lu, and Chen]{guo2024prediction}
Y.~Guo, Y.~Hu, J.~Zhang, Y.-J. Wang, X.~Chen, C.~Lu, and J.~Chen.
\newblock Prediction with action: Visual policy learning via joint denoising process.
\newblock In \emph{The Thirty-eighth Annual Conference on Neural Information Processing Systems}, 2024.

\bibitem[Li et~al.(2025)Li, Gao, Sadigh, and Song]{li2025unified}
S.~Li, Y.~Gao, D.~Sadigh, and S.~Song.
\newblock Unified video action model.
\newblock \emph{arXiv preprint arXiv:2503.00200}, 2025.

\bibitem[Zhu et~al.(2025)Zhu, Yu, Feng, Burchfiel, Shah, and Gupta]{zhu2025unified}
C.~Zhu, R.~Yu, S.~Feng, B.~Burchfiel, P.~Shah, and A.~Gupta.
\newblock Unified world models: Coupling video and action diffusion for pretraining on large robotic datasets.
\newblock \emph{arXiv preprint arXiv:2504.02792}, 2025.

\bibitem[Zhao et~al.(2025)Zhao, Lu, Kim, Fu, Zhang, Wu, Li, Ma, Han, Finn, et~al.]{zhao2025cot}
Q.~Zhao, Y.~Lu, M.~J. Kim, Z.~Fu, Z.~Zhang, Y.~Wu, Z.~Li, Q.~Ma, S.~Han, C.~Finn, et~al.
\newblock Cot-vla: Visual chain-of-thought reasoning for vision-language-action models.
\newblock \emph{arXiv preprint arXiv:2503.22020}, 2025.

\bibitem[McCarthy et~al.(2024)McCarthy, Tan, Schmidt, Acero, Herr, Du, Thuruthel, and Li]{mccarthy2024generalistrobotlearninginternet}
R.~McCarthy, D.~C. Tan, D.~Schmidt, F.~Acero, N.~Herr, Y.~Du, T.~G. Thuruthel, and Z.~Li.
\newblock Towards generalist robot learning from internet video: A survey.
\newblock \emph{arXiv preprint arXiv:2404.19664}, 2024.

\bibitem[Grauman et~al.(2022)Grauman, Westbury, Byrne, Chavis, Furnari, Girdhar, Hamburger, Jiang, Liu, Liu, et~al.]{grauman2022ego4d}
K.~Grauman, A.~Westbury, E.~Byrne, Z.~Chavis, A.~Furnari, R.~Girdhar, J.~Hamburger, H.~Jiang, M.~Liu, X.~Liu, et~al.
\newblock Ego4d: Around the world in 3,000 hours of egocentric video.
\newblock In \emph{Proceedings of the IEEE/CVF Conference on Computer Vision and Pattern Recognition}, 2022.

\bibitem[Nair et~al.(2022)Nair, Rajeswaran, Kumar, Finn, and Gupta]{nair2022r3m}
S.~Nair, A.~Rajeswaran, V.~Kumar, C.~Finn, and A.~Gupta.
\newblock R3m: A universal visual representation for robot manipulation.
\newblock \emph{arXiv preprint arXiv:2203.12601}, 2022.

\bibitem[Dasari et~al.(2023)Dasari, Srirama, Jain, and Gupta]{dasari2023unbiased}
S.~Dasari, M.~K. Srirama, U.~Jain, and A.~Gupta.
\newblock An unbiased look at datasets for visuo-motor pre-training.
\newblock In \emph{Conference on Robot Learning}, 2023.

\bibitem[Zeng et~al.(2024)Zeng, Bu, Wang, Xia, Chen, Dong, Song, Wang, Hu, Luo, et~al.]{zeng2024learning}
J.~Zeng, Q.~Bu, B.~Wang, W.~Xia, L.~Chen, H.~Dong, H.~Song, D.~Wang, D.~Hu, P.~Luo, et~al.
\newblock Learning manipulation by predicting interaction.
\newblock \emph{arXiv preprint arXiv:2406.00439}, 2024.

\bibitem[Bahl et~al.(2023)Bahl, Mendonca, Chen, Jain, and Pathak]{bahl2023affordances}
S.~Bahl, R.~Mendonca, L.~Chen, U.~Jain, and D.~Pathak.
\newblock Affordances from human videos as a versatile representation for robotics.
\newblock In \emph{Proceedings of the IEEE/CVF Conference on Computer Vision and Pattern Recognition}, 2023.

\bibitem[Kannan et~al.(2023)Kannan, Shaw, Bahl, Mannam, and Pathak]{kannan2023deft}
A.~Kannan, K.~Shaw, S.~Bahl, P.~Mannam, and D.~Pathak.
\newblock Deft: Dexterous fine-tuning for real-world hand policies.
\newblock \emph{arXiv preprint arXiv:2310.19797}, 2023.

\bibitem[Srirama et~al.(2024)Srirama, Dasari, Bahl, and Gupta]{srirama2024hrp}
M.~K. Srirama, S.~Dasari, S.~Bahl, and A.~Gupta.
\newblock Hrp: Human affordances for robotic pre-training.
\newblock \emph{arXiv preprint arXiv:2407.18911}, 2024.

\bibitem[Shaw et~al.(2023)Shaw, Bahl, and Pathak]{shaw2023videodex}
K.~Shaw, S.~Bahl, and D.~Pathak.
\newblock Videodex: Learning dexterity from internet videos.
\newblock In \emph{Conference on Robot Learning}, 2023.

\bibitem[Wen et~al.(2023)Wen, Lin, So, Chen, Dou, Gao, and Abbeel]{wen2023any}
C.~Wen, X.~Lin, J.~So, K.~Chen, Q.~Dou, Y.~Gao, and P.~Abbeel.
\newblock Any-point trajectory modeling for policy learning.
\newblock \emph{arXiv preprint arXiv:2401.00025}, 2023.

\bibitem[Bharadhwaj et~al.(2024)Bharadhwaj, Mottaghi, Gupta, and Tulsiani]{bharadhwaj2024track2act}
H.~Bharadhwaj, R.~Mottaghi, A.~Gupta, and S.~Tulsiani.
\newblock Track2act: Predicting point tracks from internet videos enables diverse zero-shot robot manipulation.
\newblock \emph{arXiv preprint arXiv:2405.01527}, 2024.

\bibitem[Wang et~al.(2023)Wang, Fan, Sun, Zhang, Fei-Fei, Xu, Zhu, and Anandkumar]{wang2023mimicplay}
C.~Wang, L.~Fan, J.~Sun, R.~Zhang, L.~Fei-Fei, D.~Xu, Y.~Zhu, and A.~Anandkumar.
\newblock Mimicplay: Long-horizon imitation learning by watching human play.
\newblock \emph{arXiv preprint arXiv:2302.12422}, 2023.

\bibitem[Zhu et~al.(2024)Zhu, Lim, Stone, and Zhu]{zhu2024vision}
Y.~Zhu, A.~Lim, P.~Stone, and Y.~Zhu.
\newblock Vision-based manipulation from single human video with open-world object graphs.
\newblock \emph{arXiv preprint arXiv:2405.20321}, 2024.

\bibitem[Bharadhwaj et~al.(2023)Bharadhwaj, Gupta, Tulsiani, and Kumar]{bharadhwaj2023zero}
H.~Bharadhwaj, A.~Gupta, S.~Tulsiani, and V.~Kumar.
\newblock Zero-shot robot manipulation from passive human videos.
\newblock \emph{arXiv preprint arXiv:2302.02011}, 2023.

\bibitem[Ye et~al.(2023)Ye, Wang, Huang, Qin, and Wang]{ye2023learning}
J.~Ye, J.~Wang, B.~Huang, Y.~Qin, and X.~Wang.
\newblock Learning continuous grasping function with a dexterous hand from human demonstrations.
\newblock \emph{IEEE Robotics and Automation Letters}, 8\penalty0 (5):\penalty0 2882--2889, 2023.

\bibitem[Qin et~al.(2022)Qin, Wu, Liu, Jiang, Yang, Fu, and Wang]{qin2022dexmv}
Y.~Qin, Y.-H. Wu, S.~Liu, H.~Jiang, R.~Yang, Y.~Fu, and X.~Wang.
\newblock Dexmv: Imitation learning for dexterous manipulation from human videos.
\newblock In \emph{European Conference on Computer Vision}, 2022.

\bibitem[Yang et~al.(2024)Yang, Cao, Deng, Antonova, Song, and Bohg]{yang2024equibot}
J.~Yang, Z.-a. Cao, C.~Deng, R.~Antonova, S.~Song, and J.~Bohg.
\newblock Equibot: Sim (3)-equivariant diffusion policy for generalizable and data efficient learning.
\newblock \emph{arXiv preprint arXiv:2407.01479}, 2024.

\bibitem[Bruce et~al.(2024)Bruce, Dennis, Edwards, Parker-Holder, Shi, Hughes, Lai, Mavalankar, Steigerwald, Apps, Aytar, Bechtle, Behbahani, Chan, Heess, Gonzalez, Osindero, Ozair, Reed, Zhang, Zolna, Clune, de~Freitas, Singh, and Rocktäschel]{bruce2024geniegenerativeinteractiveenvironments}
J.~Bruce, M.~Dennis, A.~Edwards, J.~Parker-Holder, Y.~Shi, E.~Hughes, M.~Lai, A.~Mavalankar, R.~Steigerwald, C.~Apps, Y.~Aytar, S.~Bechtle, F.~Behbahani, S.~Chan, N.~Heess, L.~Gonzalez, S.~Osindero, S.~Ozair, S.~Reed, J.~Zhang, K.~Zolna, J.~Clune, N.~de~Freitas, S.~Singh, and T.~Rocktäschel.
\newblock Genie: Generative interactive environments, 2024.
\newblock URL \url{https://arxiv.org/abs/2402.15391}.

\bibitem[Chen et~al.(2025)Chen, Ge, Tang, Li, Ge, Ding, Shan, and Liu]{chen2025motolatentmotiontoken}
Y.~Chen, Y.~Ge, W.~Tang, Y.~Li, Y.~Ge, M.~Ding, Y.~Shan, and X.~Liu.
\newblock Moto: Latent motion token as the bridging language for learning robot manipulation from videos, 2025.
\newblock URL \url{https://arxiv.org/abs/2412.04445}.

\bibitem[Schmidt and Jiang(2024)]{schmidt2024learning}
D.~Schmidt and M.~Jiang.
\newblock Learning to act without actions.
\newblock In \emph{The Twelfth International Conference on Learning Representations}, 2024.
\newblock URL \url{https://openreview.net/forum?id=rvUq3cxpDF}.

\bibitem[Ren et~al.(2025)Ren, Wei, Guo, Zhao, Kang, Feng, and Jin]{ren2025videoworldexploringknowledgelearning}
Z.~Ren, Y.~Wei, X.~Guo, Y.~Zhao, B.~Kang, J.~Feng, and X.~Jin.
\newblock Videoworld: Exploring knowledge learning from unlabeled videos, 2025.
\newblock URL \url{https://arxiv.org/abs/2501.09781}.

\bibitem[Bu et~al.(2025)Bu, Yang, Cai, Gao, Ren, Yao, Luo, and Li]{bu2025univla}
Q.~Bu, Y.~Yang, J.~Cai, S.~Gao, G.~Ren, M.~Yao, P.~Luo, and H.~Li.
\newblock Univla: Learning to act anywhere with task-centric latent actions.
\newblock \emph{arXiv preprint arXiv:2505.06111}, 2025.

\bibitem[Gao et~al.(2025)Gao, Zhou, Du, Zhang, and Gan]{gao2025adaworld}
S.~Gao, S.~Zhou, Y.~Du, J.~Zhang, and C.~Gan.
\newblock Adaworld: Learning adaptable world models with latent actions.
\newblock \emph{arXiv preprint arXiv:2503.18938}, 2025.

\bibitem[Bansal et~al.(2024)Bansal, Bitton, Szpektor, Chang, and Grover]{bansal2024videocon}
H.~Bansal, Y.~Bitton, I.~Szpektor, K.-W. Chang, and A.~Grover.
\newblock Videocon: Robust video-language alignment via contrast captions.
\newblock In \emph{Proceedings of the IEEE/CVF Conference on Computer Vision and Pattern Recognition}, pages 13927--13937, 2024.

\bibitem[Cadene et~al.(2024)Cadene, Alibert, Soare, Gallouedec, and Wolf]{cadene2024lerobot}
R.~Cadene, S.~Alibert, A.~Soare, Q.~Gallouedec, and T.~Wolf.
\newblock Lerobot: Making ai for robotics more accessible with end-to-end learning, 2024.
\newblock URL \url{https://github.com/huggingface/lerobot}.
\newblock Accessed: 2025-04-30.

\end{thebibliography}

\clearpage

\appendix

\section{Extracting Pseudo Actions from Synthetic Videos}
\label{appen:pseudo_actions}

\begin{table}[]
\centering
\caption{\centering LAPA Training Dataset Statistics}
\begin{tabular}{lllll}
\toprule
Dataset & Length (Frames) & Duration (hr) & FPS & Category \\ 
\midrule
GR-1 Teleop Pre-Training & 6.4M & 88.4 & 20  & Real robot \\
DexMG & 4.4M & 61.64 & 20 & Simulation \\
DROID (OXE) & 23.1M & 428.3 & 15 & Real robot \\
RT-1 (OXE) & 3.7M & 338.4 & 3  & Real robot \\
Language Table (OXE) & 7.0M & 195.7 & 10 & Real robot \\
Bridge-v2 (OXE) & 2.0M & 111.1 & 5 &  Real robot \\
RoboCasa & 19.3M & 268.0 & 20 & Simulation \\
Agibot-Alpha & 213.8M & 1,979.4 & 30 & Real robot \\
Sth-v2 & 4.0M & 105.7 & 30 & Human \\
Ego4D & 154.4M & 2,144.7 & 20  & Human \\
\midrule
Total & 438.1M & 5,721.3 & -- & --  \\   
\bottomrule
\end{tabular}
\label{tab:dataset_stats}
\end{table}

\begin{figure}[t!]
    \includegraphics[width=0.99\columnwidth]{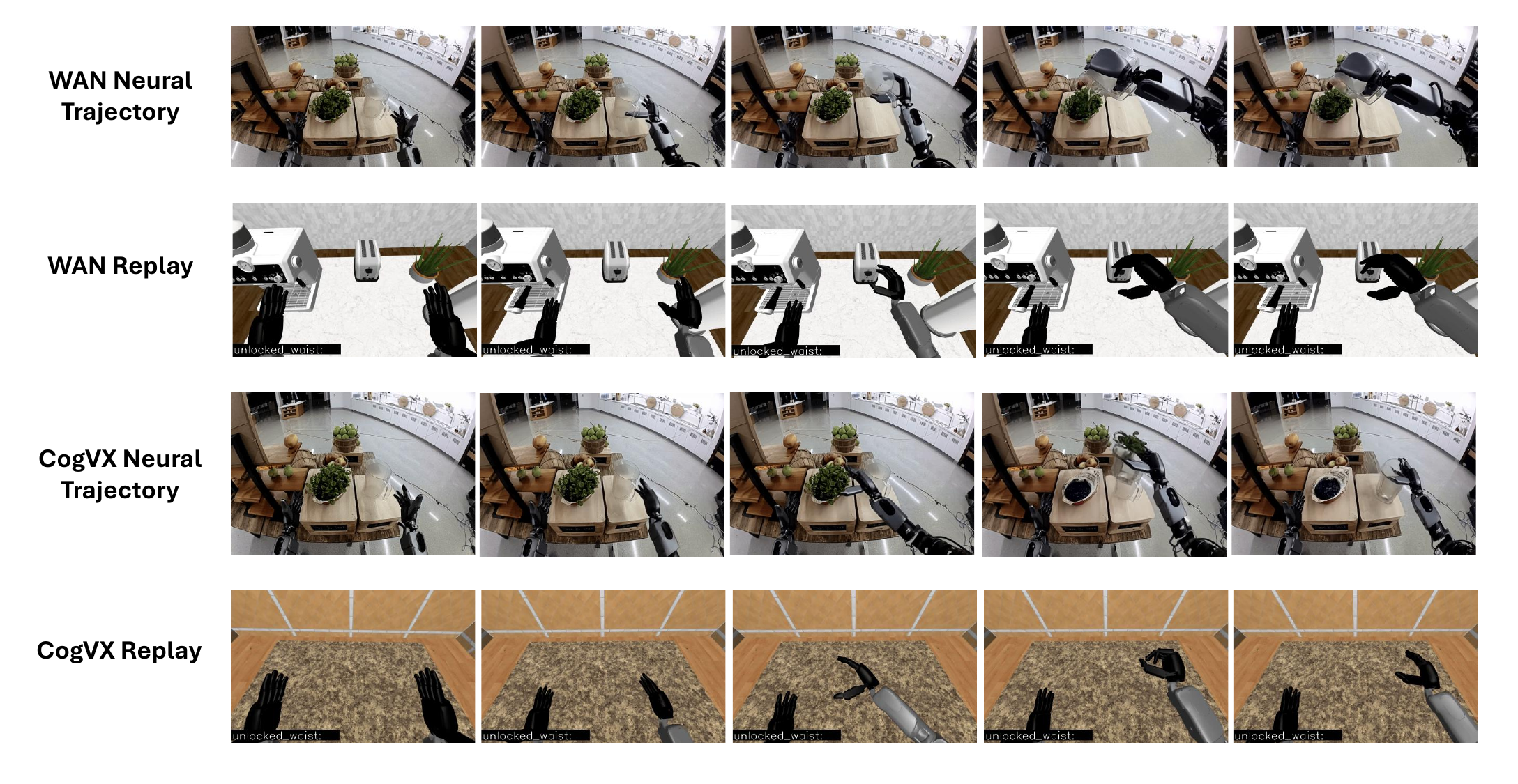}
    \caption{\textbf{Neural Trajectories and Replay Videos for WAN and CogVideoX model}. The language instruction is to ``Use the right hand to pick up the plastic pitcher and pour water onto the green plant.''}
    \label{fig:idm_replay}
\end{figure}
For IDM, if we have a digital cousin of the real robot embodiment in simulation, we can also replay the pseudo actions in simulation and do intermediate checking whether the neural trajectory quality is not good enough or the bottleneck is on the IDM model (as shown in Figure \ref{fig:idm_replay}). Empirically, we observe that most of the bottleneck is from the quality of the neural trajectories, which indicates that future video models that can generate videos with better language following and physics alignment could lead to a significant boost on the downstream task. For LAPA training, we trained a collection of datasets that include real robots, simulation, and human videos. The detailed statistics are shown in Table \ref{tab:dataset_stats}. We use a codebook size of 8 and a sequence length of 16 for vector quantization. We train 100K steps with a batch size of 1024.

\section{Environment for Teleoperation and Evaluation}
\label{appen:env}

\begin{figure}[t!]
    \includegraphics[width=0.99\columnwidth]{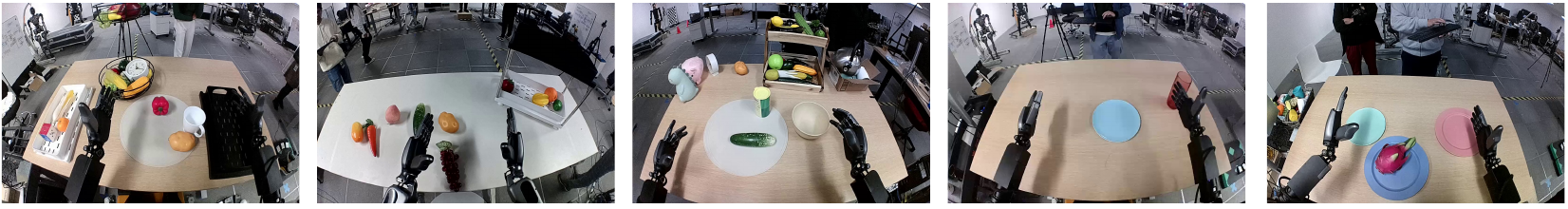}
    \caption{\textbf{Seen Environment.} Sample images for the environment where we collected the pick-and-place GR1 data.}
    \label{fig:gr1_env}
\end{figure}

\begin{figure}[t!]
    \includegraphics[width=0.99\columnwidth]{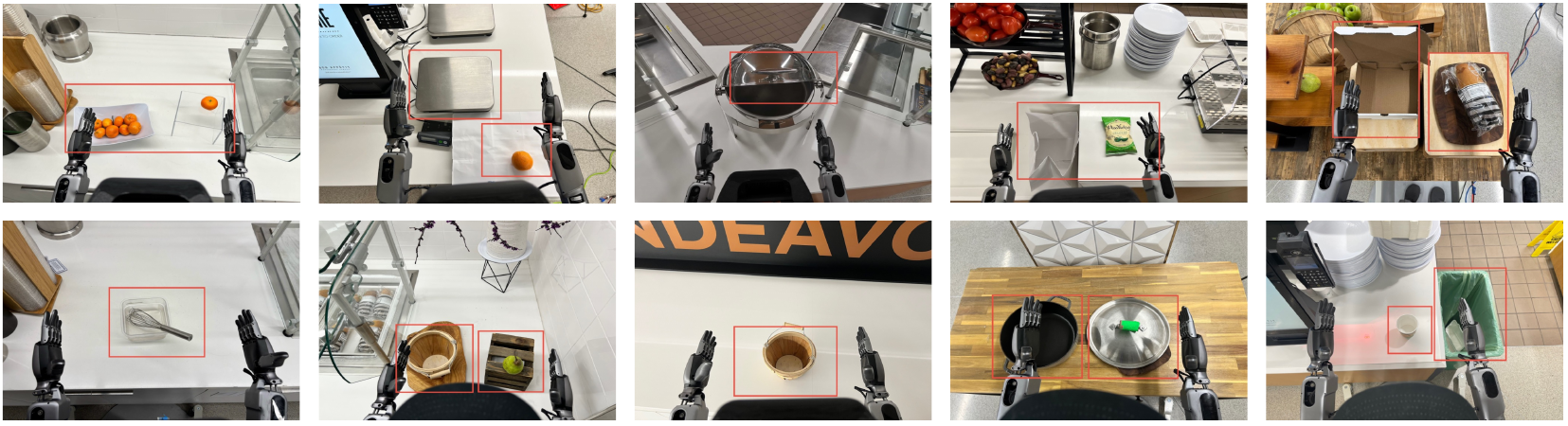}
    \caption{\textbf{Unseen Environment.} All of the 10 environments for our environment generalization experiments.}
    \label{fig:gen_env}
\end{figure}

 We provide some sample images of the environment where we collected all of our GR1 humanoid teleoperation data in Figure \ref{fig:gr1_env} and all of the 10 environments where we conducted environment generalization results in Figure \ref{fig:gen_env}, respectively.

\section{Examples of Multiview Robot Data Processing}
\label{appen:multiview_examples}

\begin{figure}[t!]
    \includegraphics[width=0.99\columnwidth]{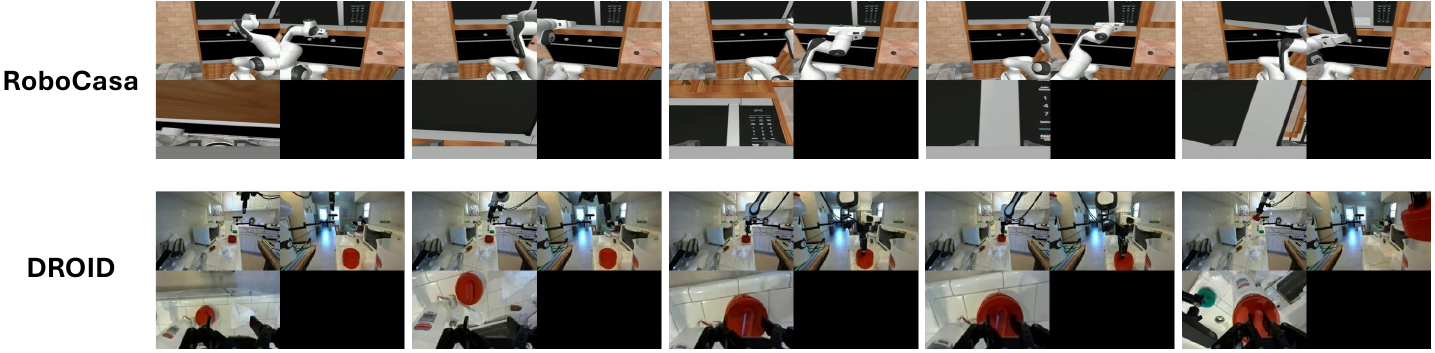}
    \caption{\textbf{Multiview Examples.} The top row shows a trajectory from RoboCasa and the bottom shows a trajectory from the DRIOD dataset.}
    \label{fig:multiview}
\end{figure}

We provide examples of how we process multiview training data, RoboCasa, and DROID, for video world model fine-tuning in Figure \ref{fig:multiview}. Specifically, we arrange the viewpoints into a 2$\times$2 grid: the left camera view is placed at the top-left, the right camera view at the top-right, and the wrist camera view at the bottom-left. A black image is inserted in the bottom-right to complete the grid.

\section{Video World Model Training Hyperparameters}
\label{appen:wm_hyperparameter}
For all of the WAN 2.1 fine-tuning experiments, we used a learning rate of 1e-4, LoRA rank 4, and LoRA alpha 4. For RoboCasa finetuning, we trained the model for 100 epochs with a batch size of 32. For GR1 finetuning, we trained the model for 75 epochs with a batch size of 64. For DROID fine-tuning, we trained the model for 5 epochs with a batch size of 64. For both of the two tasks in SO-100 finetuning, we trained the model for 200 epochs with batch size 8.

\section{Detailed Experimental Results on RoboCasa}
\label{appen:robocasa_detail_result}
Table \ref{table:robocasa_detail_result} shows all of the experimental results on RoboCasa. As seen in the chart, ONLY neural trajectories also achieves 20.55\% average success rate across the 24 tasks, showcasing how close neural trajectories are to ground truth trajectories.

\begin{table}[htbp]
\centering
\caption{\textbf{Experimental Results on RoboCasa}. \textit{NT} stands for 240k neural trajectories.}
\label{table:robocasa_detail_result}
\resizebox{\textwidth}{!}{%
\begin{tabular}{llccc|ccc|c}
\toprule
\multicolumn{2}{c}{\multirow{2}{*}{Task}} & \multicolumn{7}{c}{GR00T N1} \\
\cmidrule(r){3-9}
\multicolumn{2}{c}{} & 30 traj. & 100 traj. & 300 traj. & 30 traj. + NT & 100 traj. + NT & 300 traj. + NT & ONLY NT \\
\midrule
\multirow{8}{*}{Pick and Place} 
& PnPCabToCounter & 0.93 & 3.92 & 19.61 & 5.77 & 13.46 & 25.00 & 1.96 \\
& PnPCounterToCab & 1.85 & 6.86 & 36.27 & 3.85 & 19.23 & 50.96 & 16.67 \\
& PnPCounterToMicrowave & 0.00 & 0.00 & 12.75 & 0.00 & 9.62 & 19.23 & 0.00 \\
& PnPCounterToSink & 0.00 & 0.98 & 9.80 & 0.00 & 12.50 & 33.65 & 1.96 \\
& PnPCounterToStove & 0.00 & 0.00 & 23.53 & 0.00 & 12.50 & 42.31 & 8.82 \\
& PnPMicrowaveToCounter & 0.00 & 0.00 & 15.69 & 0.00 & 14.42 & 28.85 & 0.00 \\
& PnPSinkToCounter & 0.00 & 5.88 & 33.33 & 3.85 & 28.85 & 60.58 & 0.98 \\
& PnPStoveToCounter & 0.00 & 0.00 & 29.41 & 0.96 & 9.62 & 58.65 & 5.88 \\
\midrule
\multirow{4}{*}{Open/Close Doors} 
& CloseDoubleDoor & 0.00 & 43.14 & 74.51 & 9.62 & 52.88 & 82.69 & 2.94 \\
& OpenDoubleDoor & 0.00 & 12.75 & 14.71 & 0.00 & 8.65 & 28.85 & 0.00 \\
& CloseSingleDoor & 49.07 & 67.65 & 83.33 & 51.92 & 80.77 & 94.23 & 52.94 \\
& OpenSingleDoor & 20.37 & 54.90 & 58.82 & 44.23 & 55.77 & 47.12 & 15.69 \\
\midrule
\multirow{2}{*}{Open/Close Drawers} 
& CloseDrawer & 76.85 & 96.08 & 99.02 & 88.46 & 98.08 & 98.08 & 82.35 \\
& OpenDrawer & 9.26 & 42.16 & 79.41 & 33.65 & 68.27 & 74.04 & 33.33 \\
\midrule
\multirow{2}{*}{Twisting Knobs} 
& TurnOnStove & 14.81 & 25.49 & 55.88 & 21.15 & 27.88 & 51.92 & 17.65 \\
& TurnOffStove & 4.63 & 15.69 & 26.47 & 7.69 & 13.46 & 25.96 & 6.86 \\
\midrule
\multirow{3}{*}{Turning Levers} 
& TurnOffSinkFaucet & 49.07 & 67.65 & 72.55 & 51.92 & 69.23 & 95.19 & 59.80 \\
& TurnSinkSpout & 24.07 & 42.16 & 52.94 & 37.50 & 45.19 & 59.62 & 28.43 \\
& TurnOnSinkFaucet & 33.33 & 59.80 & 62.75 & 48.08 & 67.31 & 72.12 & 25.49 \\
\midrule
\multirow{3}{*}{Pressing Buttons} 
& TurnOffMicrowave & 47.22 & 57.84 & 70.59 & 55.77 & 75.96 & 76.92 & 29.41 \\
& TurnOnMicrowave & 55.56 & 73.53 & 78.43 & 49.04 & 52.88 & 72.12 & 48.04 \\
& CoffeePressButton & 27.78 & 56.86 & 85.29 & 34.62 & 63.46 & 83.65 & 48.04 \\
\midrule
\multirow{2}{*}{Insertion} 
& CoffeeServeMug & 3.70 & 34.31 & 72.55 & 11.54 & 48.08 & 74.04 & 2.94 \\
& CoffeeSetupMug & 0.00 & 1.96 & 22.55 & 0.00 & 10.58 & 26.92 & 2.94 \\
\midrule
\multicolumn{2}{c}{Average} & 17.44 & 32.07 & 49.59 & 23.32 & 39.94 & 57.61 & 20.55 \\
\bottomrule
\end{tabular}
}
\end{table}

\section{Fine-tuning Data for Video World Models and IDMs}
\label{appen:video_world_model}
In this section, we provide some detailed information about the protocol we followed to train the video world models and the IDM for each experimental setup.

\paragraph{Four dexterous tasks on Real-world GR1.} To train our video world model, we follow the same protocol outlined in Section \ref{sec:method}, and train on 2,884 GR1 trajectories of pick-and-place collected in a single lab environment. Since these four tasks differ significantly from the target task, we further fine-tune the model on the \textit{low data} trajectories for each task. For each task, we collect 100 trajectories, but only utilize 10 trajectories for Hammering, Wiping, Stacking, and 25 trajectories for Folding to test data efficiency. We utilize the IDM trained only on the 2,884 GR1 pick-and-place data for all experimentsl.

\paragraph{3 tasks on Franka.}
Following protocol in Section \ref{sec:method}, we train our video world model on 49,895 DROID data examples, and further fine-tune the model on the \textit{low data} trajectories for each task. We found that utilizing the model trained only from the DROID dataset results in dreams that show generalization to the new environment, but produced trajectories that made mistakes on fine-graed details (e.g. grasping). We use 11, 10, and 8 trajectories for putting milk in bowl, cube stacking, and scooping M\&Ms, respectively. Similarly to GR1, we use the IDM trained on 49,895 trajectories and do not do any specific post-training.

\paragraph{2 tasks on SO-100.} The original SO-100 videos concatenate multiple trajectories with identical actions into a single video. For fine-tuning, we manually trim and split these into separate videos, each corresponding to an individual trajectory. Specifically, we sample 10 and 13 videos for the two tasks, which yield 68 and 44 trajectories, respectively, after trimming.

\section{Full Real-world Experimental Results}

\begin{table}[htbp]
\centering
\caption{\textbf{Success Rate (\%) of Real-world Data Augmentation Experiments.}.} 
\resizebox{\textwidth}{!}{%
\begin{tabular}{lcccccccccc}
\toprule
\multicolumn{1}{l}{\multirow{2}{*}{Model}} & \multicolumn{5}{c}{GR1} & \multicolumn{3}{c}{Franka} & \multicolumn{2}{c}{SO-100} \\
\cmidrule(r){2-6} \cmidrule(r){7-9} \cmidrule(r){10-11}
\multicolumn{1}{c}{} & Hammering & Wiping & Folding & Stacking & Average & Pick\&Place & Cube Stacking & Tool Usage & Pick\&Place & Tic-Tac-Toe \\
\midrule
DP & 35.0 & 23.3 & 6.6 & 25.0 & 22.0 & 20.0 & 0.0 & 10.0 & - & - \\
$\pi_0$ & - & - & - & - & - & 30.0 & 10.0 & 20.0 & - & - \\
GR00T N1 & 60.0 & 36.6 & 27.0 & 25.0 & 37.0 & 40.0 & 10.0 & 20.0 & 17.0 & 25.0 \\
\midrule
DP + Neural Traj. & 15.0 & 33.3 & 26.4 & 35.0 & 27.0 & 30.0 & 20.0 & 10.0 & - & - \\
$\pi_0$ + Neural Traj. & - & - & - & - & - & 40.0 & 20.0 & 20.0 & - & - \\
GR00T N1 + Neural Traj. & 65.0 & 49.0 & 37.0 & 35.0 & 46.0 & 60.0 & 20.0 & 30.0 & 26.0 & 65.0 \\
\midrule
DP (High Data) & 60.0 & 36.0 & 43.3 & 75.0 & 54.0 & 30.0 & 20.0 & 20.0 & - & - \\
$\pi_0$ (High Data) & - & - & - & - & - & 50.0 & 40.0 & 40.0 & - & - \\
GR00T N1 (High Data) & 75.0 & 50.0 & 66.6 & 85.0 & 69.0 & 80.0 & 50.0 & 40.0 & 36.0 & 40.0 \\
\bottomrule
\end{tabular}
}
\label{tab:robotics_performance}
\end{table}

\label{appen:full_realworld_results}
Table \ref{tab:robotics_performance} shows the entire experimental results, including the model performance when trained on the ``High Data'' variant of each experimental setup. 

\section{Video World Model Evaluation}

\subsection{Success Rate}
\label{sec:sr}
Specifically, we use the following prompts to Qwen2.5-VL-7B-Instruct~\citep{Qwen2.5-VL} to judge whether a video follows the instruction to complete a specific task or not.

\begin{tcolorbox}[title=Prompt Template for Success Rate]
\textbf{User:} \{Video: \texttt{<vid\_path>}\}\{Text: "The video shows a robot arm completing a specific task. Please evaluate: if the video follows the instruction to finish the task '\{prompt\}', give a positive score. Reply only '0' for No or '1' for Yes."\}

\textbf{Assistant:} 0 or 1
\end{tcolorbox}

\begin{tcolorbox}[title=Prompt Template for Success Rate (Zeroshot)]
\textbf{User:} \{Video: \texttt{<vid\_path>}\}\{Text: "You are evaluating if a robot arm correctly follows this instruction: '\{prompt\}'

CRITICAL EVALUATION PROCESS:
1. FIRST CHECK: If you see HUMAN HANDS instead of robot arms, IMMEDIATELY ANSWER 0.
2. SECOND CHECK: Only if robot arms confirmed, verify if the instruction is followed exactly.
3. For videos with multiview clip (4 grids), verify if the instruction is followed exactly in each view. Only if all the view is following instruction, answer 1, otherwise, answer 0.

Remember: human hands = automatic failure (0). Be extremely strict in your judgment.
For videos with multiview clip (4 grids), check if the human arm is present in any view, if so, make sure to answer 0.

Reply ONLY with a single digit: 0 for failure or 1 for success."\}

\textbf{Assistant:} 0 or 1
\end{tcolorbox}

\subsection{Physics Alignment}
\label{sec:pa}
While human evaluation provides accurate benchmarking, it is time-consuming and costly at scale. To enable model developers with limited resources to use our benchmark, we use \textbf{VideoCon-Physics}, an open video-text language model with $7$B parameters trained on real videos for physics alignment evaluation~\citep{bansal2024videophy}. Specifically, they finetune VideoCon~\citep{bansal2024videocon} using human annotations collected for physics alignment on generated videos. We prompt it to generate binary responses conditioned on multimodal templates. They evaluate this auto-rater by computing ROC-AUC between human judgments and model predictions on videos generated with testing prompts, and show that they have a strong correlation with human evaluation results.
In addition to it, we use Qwen2.5-VL-7B-Instruct~\citep{Qwen2.5-VL} to judge whether a video follow physics or not with the following prompt:

\begin{tcolorbox}[title=Prompt Template for Physics Alignment]
\textbf{User:} \{Video: \texttt{<vid\_path>}\}\{"The video shows a robot arm completing a specific task. Does the video show good physics dynamics that is aligned with the physical world? Answer 0 for No or 1 for Yes. Reply only 0 or 1."\}

\textbf{Assistant:} 0 or 1
\end{tcolorbox}
We finally compute the average of two scores together for each video.

\subsection{Human Evaluation}
\label{sec:human}
To verify the reliability of our automatic benchmark on success rate, we compare it with human evaluation results and calculate the AUC-ROC between them.
In detail, we perform human evaluations of all of the instances from the 3 fine-tuned video world models from Table \ref{tab:wm_results}, to show that the model-based metrics indeed do correlate with human-based judgement of success rate (SR) and physics alignment (PA). For SR, similar to the model-based metric, humans give a binary signal, 0 or 1, whether the trajectory has successfully completed the task specified by the language. For PA, instead of giving a fine-grained score, humans rank the model's output, given the same initial frame, and see the ranking corresponds to the ranking by the scores of the model.

\begin{table}[ht]
\centering
\begin{tabular}{llccccc}
\toprule
Dataset & Metric     & Hunyuan‐sft & CogVideoX‐sft & WAN2.1‐sft & Cosmos-sft & Pearson~$r$ \\
\midrule
\multirow{2}{*}{RoboCasa} 
  & IF       &  68.8 & 72.9 & 77.1 & 79.2  & \multirow{2}{*}{0.94} \\
  & IF‐human &  81.3 & 79.2 & 91.7 & 93.8  &                      \\
\midrule
\multirow{2}{*}{GR1‐Object} 
  & IF       &  38.0 & 72.0 & 72.0 & 90.0  & \multirow{2}{*}{0.93} \\
  & IF‐human &  52.0 & 72.0 & 80.0 & 84.0  &                      \\
\midrule
\multirow{2}{*}{GR1‐Behavior} 
  & IF       &  38.3 & 44.0 & 72.3 & 59.6  & \multirow{2}{*}{0.96} \\
  & IF‐human &  14.9 & 21.3 & 74.5 & 68.1  &                      \\
\midrule
\multirow{2}{*}{GR1‐Env} 
  & IF       &  27.6 & 55.2 & 48.3 & 69.0  & \multirow{2}{*}{1.00} \\
  & IF‐human &  20.0 & 30.0 & 43.3 & 53.3  &                      \\
\bottomrule
\end{tabular}
\caption{Pearson correlation coefficients between automatic IF (GPT-4o) and human IF‐human scores across different datasets and model variants.}
\label{tab:if_correlation_new}
\end{table}

\begin{table}[ht]
\centering
\begin{tabular}{llccccc}
\toprule
Dataset & Metric     & Hunyuan‐sft & CogVideoX‐sft & WAN2.1‐sft & Cosmos‐sft & Pearson~$r$ \\
\midrule
\multirow{2}{*}{RoboCasa} 
  & IF        &   8.3 & 10.4 & 18.8 & 29.2  & \multirow{2}{*}{0.92} \\
  & IF‐human  &  81.3 & 79.2 & 91.7 & 93.8  &                       \\
\midrule
\multirow{2}{*}{GR1‐Object} 
  & IF        &  26.0 & 38.0 & 58.0 & 62.0  & \multirow{2}{*}{0.95} \\
  & IF‐human  &  52.0 & 72.0 & 80.0 & 84.0  &                       \\
\midrule
\multirow{2}{*}{GR1‐Behavior} 
  & IF        &  10.6 & 28.0 & 55.3 & 61.7  & \multirow{2}{*}{0.97} \\
  & IF‐human  &  14.9 & 21.3 & 70.2 & 68.1  &                       \\
\midrule
\multirow{2}{*}{GR1‐Env} 
  & IF        &  27.6 & 41.4 & 65.5 & 65.5  & \multirow{2}{*}{0.96} \\
  & IF‐human  &  20.0 & 30.0 & 43.3 & 53.3  &                       \\
\bottomrule
\end{tabular}
\caption{Pearson correlation coefficients between automatic IF (Qwen2.5-VL) and human IF‐human scores for each dataset.}
\label{tab:if_correlation_new_qwen}
\end{table}

Table \ref{tab:if_correlation_new} and Table \ref{tab:if_correlation_new_qwen} present the Pearson correlation coefficients between our automatic evaluation metric (IF) and the corresponding human‐annotated scores (IF‐human) for three model variants on each dataset. The correlations of IF evaled by GPT-4o are uniformly high—0.94 for RoboCasa, 0.93 for GR1‐Object, 0.96 for GR1‐Behavior, and essentially 1.00 for GR1‐Env—indicating a near‐perfect linear relationship across all cases. These results confirm that the IF metric faithfully captures human judgments and can serve as a reliable proxy for resource‐intensive manual evaluation.


\subsection{Intermediary Step for Checking Downstream Performance}
\label{sec:intermediary}

The most straightforward way to truly quantify the capabilities of the video world models is to use them to generate neural trajectories and use the generated trajectories for downstream visuomotor policy training. In fact, we generate 7k neural trajectories for each of the video world models (zero-shot and fine-tuned) from Table \ref{tab:wm_results} and show that benchmark numbers directly correlate to downstream robot policy performances. However, this is very resource-intensive, since verifying a new video world model beyond benchmark numbers requires generating 7k new videos. As an intermediary step, we utilize a \textit{cheaper} way of quantifying the quality of the dreams. After extracting the IDM actions from the generated videos (see Section \ref{subsec:pseudo_actions}), we replay the IDM actions in simulation, where we have access to the digital twin of the Fourier GR1. Some examples of replayed IDM actions in simulation are shown in Appendix \ref{appen:pseudo_actions}. 

\section{Robot Experiment Evaluation}
\label{appendix:robot-evals}


%

%

%

\subsection{GR1 Humanoid Experiments}
\label{appen:gr1_eval}
\begin{figure}[h!]
\centering
\includegraphics[width=\textwidth]{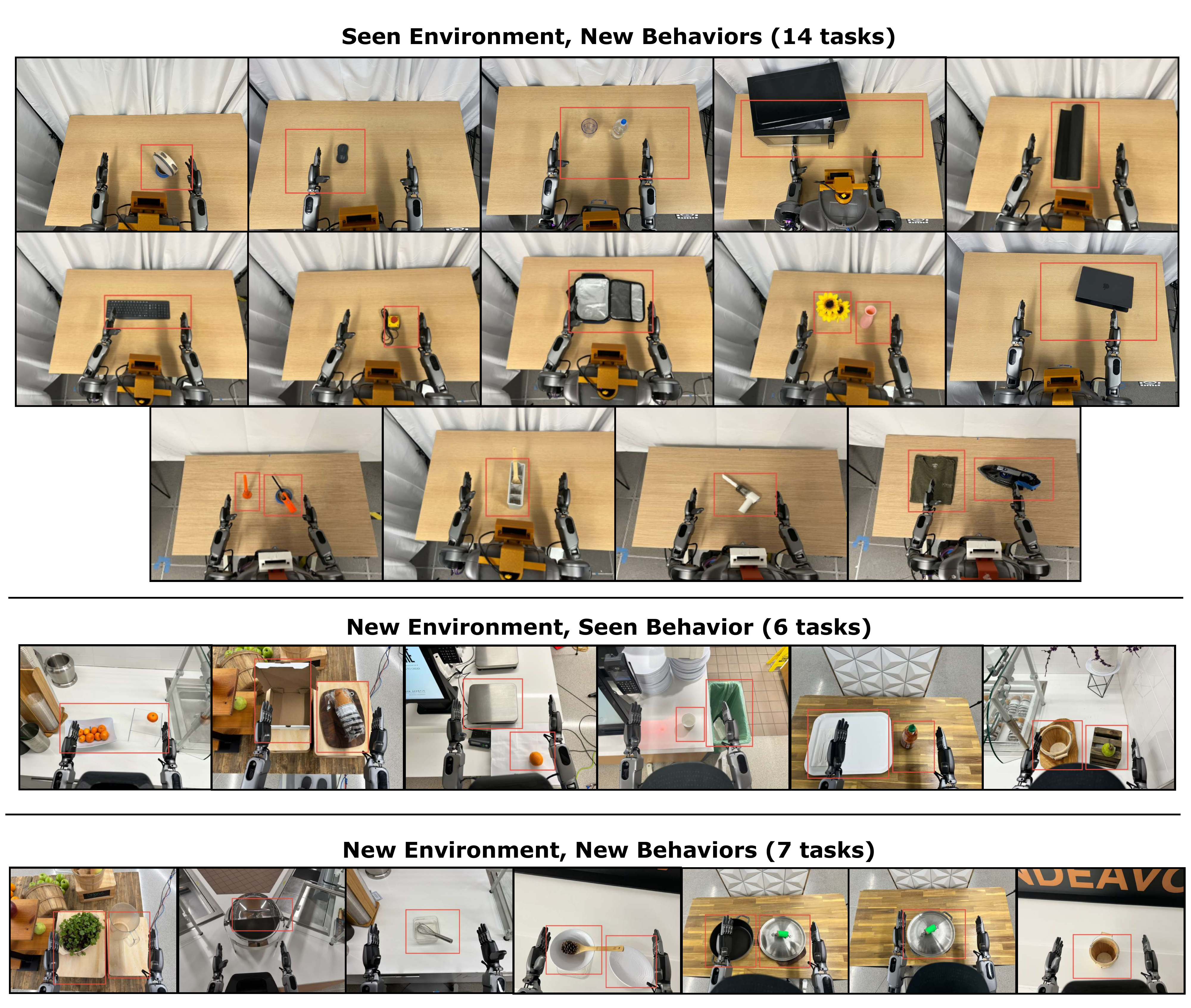}
\caption{\textbf{Evaluations for all Real-world GR1 Experiments.} The rectangular box represents the region where we randomize the target object during evaluation.}
\label{fig:eval_frames}
\end{figure}

\paragraph{Data Augmentation}
We have 4 tasks for the data augmentation experiments using the GR1 Humanoid: Hammering, Wiping, Folding, and Stacking. For each task, we collect 100 trajectories, while randomizing the target object locations in the rectangular box as shown in Figure \ref{fig:real_robot_mg}. 

For Hammering, we give 0.5 for picking up the hammer, and 1.0 for actually hitting the nail. For Wiping, 0.33 for grabbing the rag, 0.66 for taking the rag to the stain, and 1.0 for actually wiping the stain. For Folding, we give 0.33 for folding the first fold, but imperfectly, 0.66 for completing the first fold, and 1.0 for completing the second fold. Lastly, for Stacking, we give 0.5 for stacking the left bowl, and 1.0 for stacking the right bowl. We perform 10 eval rollouts per checkpoint.

\paragraph{Behavior and Environment Generalization}
Table \ref{tab:gen_criteria} shows the criterion we use to measure the performance on behavior and environment generalization. We performed 10 rollouts per checkpoint while randomizing the initial location of the target object across all trials to ensure fair, direct comparisons between models. The region of target object randomization is shown in Figure \ref{fig:eval_frames}.

\begin{table}[htbp]
\centering
\caption{\textbf{Task Evaluation Criteria for GR1 Generalization Experiments}}
\label{tab:gen_criteria}
\begin{subtable}[t]{0.48\textwidth}
\centering
\caption*{\textbf{Seen Environments, Novel Behaviors}}
\resizebox{\linewidth}{!}{%
\begin{tabular}{l l}
\toprule
\textbf{Task} & \textbf{Criteria} \\
\midrule
Open Microwave & \begin{tabular}[c]{@{}l@{}}0.33 grasp handle \\ 0.66 do closing motion \\ 1.0 close microwave\end{tabular} \\
\midrule
Open Macbook & \begin{tabular}[c]{@{}l@{}}0.5 opening motion \\ 1.0 open laptop\end{tabular} \\
\midrule
Close Lunchbox & \begin{tabular}[c]{@{}l@{}}0.5 contact lid \\ 1.0 close lunchbox\end{tabular} \\
\midrule
Hit Tambourine & \begin{tabular}[c]{@{}l@{}}0.5 grab tambourine \\ 1.0 hit with left hand\end{tabular} \\
\midrule
Hit Keyboard & \begin{tabular}[c]{@{}l@{}}0.5 going to keyboard \\ 1.0 pressing\end{tabular} \\
\midrule
Grab Button & \begin{tabular}[c]{@{}l@{}}0.5 go to button \\ 1.0 grab button\end{tabular} \\
\midrule
Pour Water & \begin{tabular}[c]{@{}l@{}}0.5 picking up \\ 1.0 pouring\end{tabular} \\
\midrule
Water Flowers & \begin{tabular}[c]{@{}l@{}}0.5 grasp pink bottle \\ 1.0 pour\end{tabular} \\
\midrule
Light Candle & \begin{tabular}[c]{@{}l@{}}0.5 grasp lighter \\ 1.0 approach candle\end{tabular} \\
\midrule
Use Vacuum & \begin{tabular}[c]{@{}l@{}}0.5 pick up vacuum \\ 1.0 do sweeping motion\end{tabular} \\
\midrule
Iron Shirt & \begin{tabular}[c]{@{}l@{}}0.5 grasp iron \\ 1.0 press shirt\end{tabular} \\
\midrule
Take Spoon Out & \begin{tabular}[c]{@{}l@{}}0.33 grasp spoon \\ 0.66 pick up spoon \\ 1.0 place spoon\end{tabular} \\
\midrule
Unroll Mat & \begin{tabular}[c]{@{}l@{}}0.5 go to mat \\ 1.0 unroll\end{tabular} \\
\midrule
Move Mouse & \begin{tabular}[c]{@{}l@{}}0.5 grab the mouse \\ 1.0 move it around\end{tabular} \\
\bottomrule
\end{tabular}%
}
\end{subtable}
\hfill
\begin{subtable}[t]{0.48\textwidth}
\centering
\caption*{\textbf{Novel Environments}}
\resizebox{\linewidth}{!}{%
\begin{tabular}{l l}
\toprule
\multicolumn{2}{c}{\textbf{Seen Behaviors}} \\
\midrule
\textbf{Task} & \textbf{Criteria} \\
\midrule
Pick up Tangerine & \begin{tabular}[c]{@{}l@{}}0.5 pick up \\ 1.0 place in bowl\end{tabular} \\
\midrule
Box Sandwich & \begin{tabular}[c]{@{}l@{}}0.5 grab the sandwich \\ 1.0 place in box\end{tabular} \\
\midrule
Weigh the Orange & \begin{tabular}[c]{@{}l@{}}0.5 pick up \\ 1.0 place on scale\end{tabular} \\
\midrule
Put Cup in Trash & \begin{tabular}[c]{@{}l@{}}0.5 grab cup \\ 1.0 throw it away\end{tabular} \\
\midrule
Put Pear in Basket & \begin{tabular}[c]{@{}l@{}}0.5 grab pear \\ 1.0 put in bucket\end{tabular} \\
\midrule
Put Sauce on Tray & \begin{tabular}[c]{@{}l@{}}0.5 grab bottle \\ 1.0 place bottle on tray\end{tabular} \\
\midrule
\multicolumn{2}{c}{\textbf{Novel Behaviors}} \\
\midrule
\textbf{Task} & \textbf{Criteria} \\
\midrule
Water Flowers & \begin{tabular}[c]{@{}l@{}}0.5 pick up pitcher \\ 1.0 water the plants\end{tabular} \\
\midrule
Lift Basket & \begin{tabular}[c]{@{}l@{}}0.5 grab handle \\ 1.0 lift bucket\end{tabular} \\
\midrule
Swirl Around Spoon & \begin{tabular}[c]{@{}l@{}}0.5 grab spoon \\ 1.0 scoop to plate\end{tabular} \\
\midrule
Use Whisk & \begin{tabular}[c]{@{}l@{}}0.5 grab whisk \\ 1.0 mix\end{tabular} \\
\midrule
Close Soup Container & \begin{tabular}[c]{@{}l@{}}0.5 use handle \\ 1.0 close\end{tabular} \\
\midrule
Uncover Pot & \begin{tabular}[c]{@{}l@{}}0.5 grab cover \\ 1.0 uncover pot\end{tabular} \\
\midrule
Cover Pot & \begin{tabular}[c]{@{}l@{}}0.5 grab cover \\ 1.0 cover pot\end{tabular} \\
\bottomrule
\end{tabular}%
}
\end{subtable}
\end{table}


\subsection{DROID (Franka) Experiments}
We carry out our second real-world study on the Franka Emika Panda arm, collecting 100 teleoperation data for three manipulation tasks, pick-and-place, cube stacking, and tool use (Figure \ref{fig:real_robot_mg}. ). We also have a \textit{low}-data regime, where we only train on 10 trajectories, except for the folding task, where we train on 25 trajectories. Following our proposed pipeline, we train our video world model and the IDM model on the DROID dataset~\citep{khazatsky2024droid}, 

To ensure rigorous evaluation, we executed 10 rollouts per checkpoint for each model and enforced identical initial state configurations across models, enabling fair, head-to-head comparisons. Within each batch of rollouts, we further randomized object poses to probe policy robustness. Results show that conditioning on neural trajectories consistently boosts the performance of Diffusion Policy, $\pi_0$, and GR00T N1 across all tasks.

\subsection{SO-100 Experiments}
We also present fine-tuning experiments with real and neural trajectories on a LeRobot SO-100 \citep{cadene2024lerobot}, serving as a new embodiment with a foundation robot policy (GR00T N1 VLA). The first task, "Picking 3 Strawberries," consists of 10 real-world trajectories and 30 neural trajectories. The second task is "Tic-Tac-Toe", which requires the correct language prompt to execute the task, and includes 13 real-world trajectories and 40 neural trajectories.

For the "Picking 3 Strawberries" task, the evaluation criteria involve 10 trials. The goal of each trial is to pick up all three strawberries from various locations on the table and place them on the plate. Each trial lasts 1 minute, with each successful pick and place contributing 33\% to the score for that trial. To ensure randomness, strawberries are placed on the left, center, and right sides of the table. In the ``Tic-Tac-Toe'' task, we evaluated the policy by prompting it with 5 tasks, each corresponding to placing an "X" in different boxes on the grid. With a total of 10 trials, the grid is randomized with varying "X" and "O" placements across the trials, each lasting 1 minute. Each successful pick and place corresponds to 0.5 points.

We observed that with co-training using neural trajectories, the policy overfits less to the proprioceptive states and conditions more effectively to the current visual state of the environment. Additionally, we noticed that the policy augmented with neural trajectories is less likely to get stuck at the initial home position, which is a common failure case of our baseline policy. Detailed results are shown in Figure \ref{fig:real_robot_mg}.

\section{Examples of Generated Neural Trajectories}
\label{appen:gen_example}
\begin{figure}[h!]
\centering
\includegraphics[width=\textwidth]{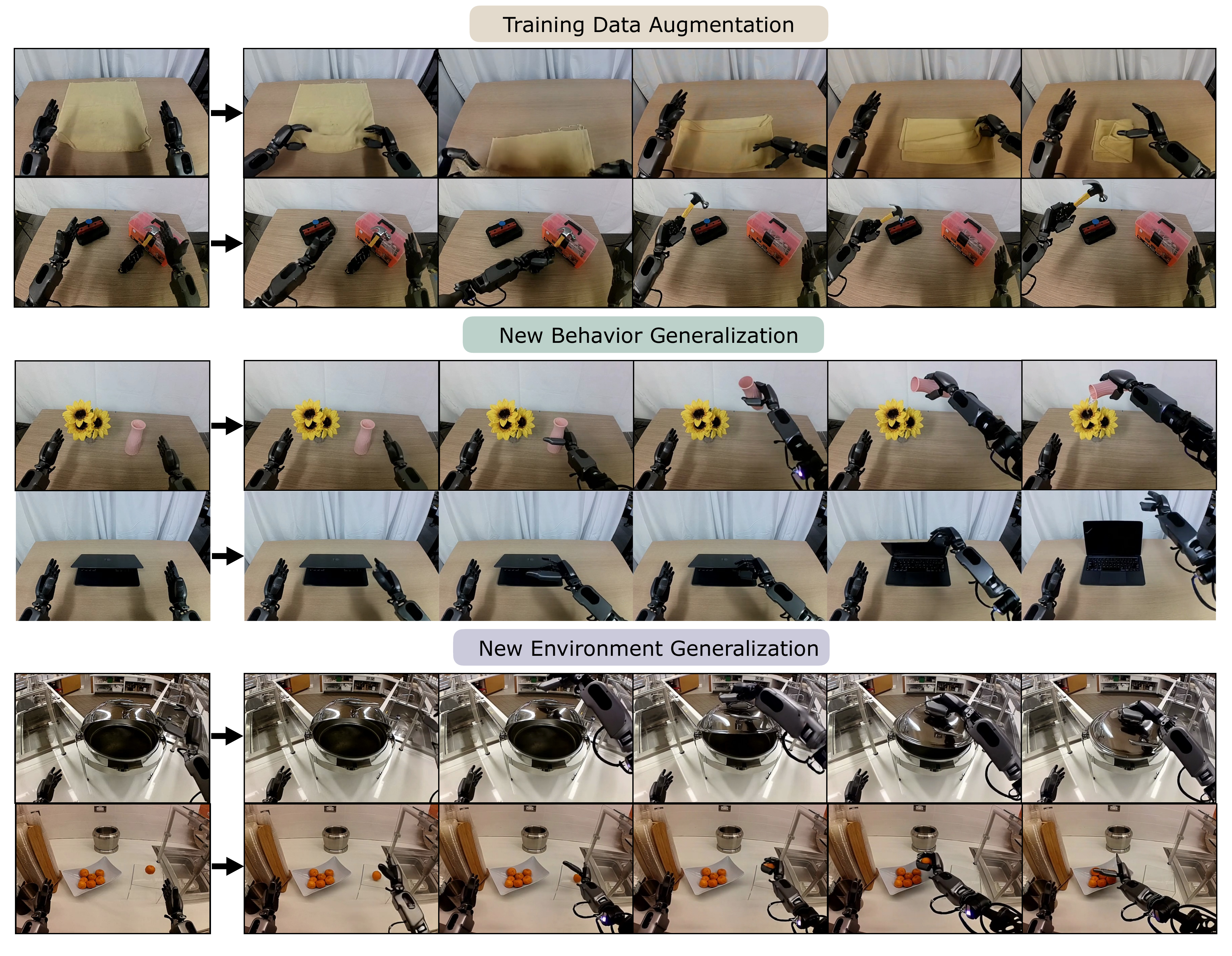}
\caption{\textbf{Examples of Neural Trajectories.}}
\label{fig:sample_dreams}
\end{figure}

\end{document}